%% file: main.tex
\documentclass[runningheads]{llncs}

\usepackage{eccv}

\usepackage{eccvabbrv}

\usepackage{graphicx}
\usepackage{booktabs}

\usepackage[accsupp]{axessibility}  % Improves PDF readability for those with disabilities.

\usepackage[pagebackref,breaklinks,colorlinks,citecolor=eccvblue]{hyperref}
\usepackage{orcidlink}

\usepackage{colortbl}
\usepackage{xcolor}
\definecolor{best}{HTML}{E8F4FD}      % 연한 파랑 (best)
\definecolor{second}{HTML}{FFF3E0}    % 연한 주황 (second-best)
\definecolor{oursrow}{HTML}{E8F4FD}   % ours row 전체 하이라이트
\usepackage{wrapfig} % preamble에 추가
\usepackage{makecell}
\usepackage{pifont}
\usepackage{multirow}
\definecolor{subheader}{HTML}{FCE4EC}  % preamble에 추가 - 연한 핑크
\definecolor{subheader_as}{HTML}{E8F5E9}  % preamble에 추가 - 연한 초록
\usepackage{float}
\usepackage{url}
\usepackage{hyperref}

\usepackage{graphicx}
\usepackage{booktabs}
\usepackage{tabularx}
\usepackage{array}
\usepackage{tcolorbox}
\tcbuselibrary{skins}
\usepackage{enumitem}

\definecolor{tocnavy}{RGB}{25, 55, 109}
\definecolor{toclight}{RGB}{219, 229, 245}
\definecolor{tocrow}{RGB}{249, 250, 252}
\definecolor{accentblue}{RGB}{25, 55, 109}
\definecolor{rowgray}{RGB}{247, 248, 250}
\definecolor{accentblue}{RGB}{25, 55, 109}

\begin{document}

\title{InSpace: Structure-Aware 3D Indoor Scene Generation from a Single 360° Image} 
% Equirectangular Image, 360° Image

\titlerunning{InSpace}

\author{
  Gwanhyeong Koo\textsuperscript{*1,2}\orcidlink{0009-0005-6455-3223} \and
  Hyunsu Kim\textsuperscript{2}\orcidlink{0000-0002-9892-2892} \and
  Youngji Kim\textsuperscript{2}\orcidlink{0000-0003-0497-7401} \and
  Taejae Lee\textsuperscript{2}\orcidlink{0009-0006-3377-8187} \and \\
  Siwoo Lim\textsuperscript{1}\orcidlink{0009-0001-1187-6548} \and
  Sunjae Yoon\textsuperscript{3}\orcidlink{0000-0001-7458-5273} \and
  Suyong Yeon\textsuperscript{$\dagger$2}\orcidlink{0009-0000-4765-4544
} \and
  Chang D. Yoo\textsuperscript{$\dagger$1}\orcidlink{0000-0002-0756-7179}
}

\authorrunning{G.~Koo et al.}

\institute{
  \textsuperscript{1}KAIST \quad
  \textsuperscript{2}NAVER LABS \quad
  \textsuperscript{3}Chung-Ang University \\
  \email{\{kookie, cd\_yoo\}@kaist.ac.kr} \
  \email{\{hyunsu.xr, suyong.yeon\}@naverlabs.com}
}

\maketitle
\renewcommand{\thefootnote}{\fnsymbol{footnote}}
\footnotetext[0]{\scriptsize\textsuperscript{*}Work done during an internship at NAVER LABS. \quad \textsuperscript{$\dagger$}Co-corresponding authors}

\vskip -0.3in
\begin{figure*}[htb]
% \captionsetup{skip=5pt}
\begin{center}
\centerline{\includegraphics[width=\textwidth]{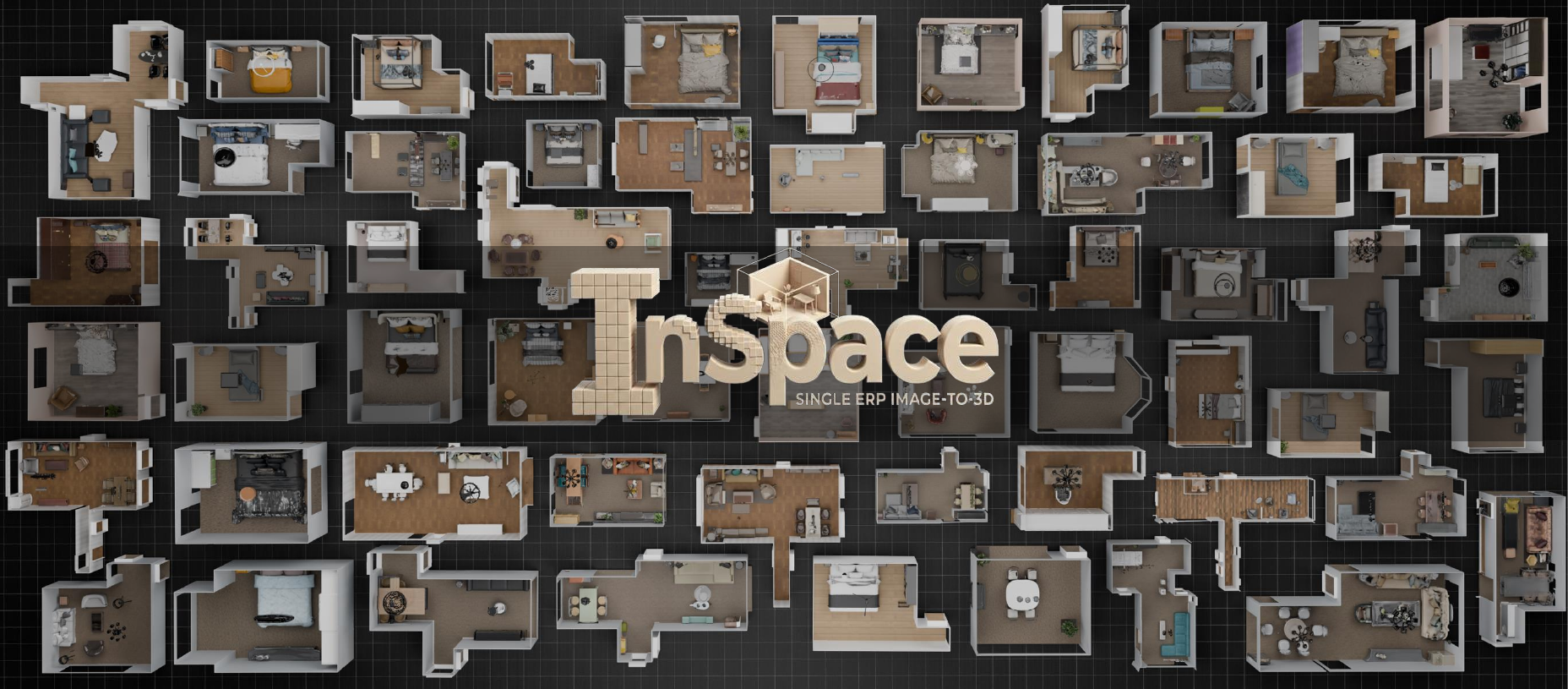}}
\vskip -0.1in
\caption{InSpace generates complete 3D indoor scenes with structural layout and individual assets from a single 360° image. Leveraging the full 360° context of panoramic images, it produces spatially coherent scenes across diverse room layouts.}
\label{fig:0_teaser}
\end{center}
\vskip -0.65in
\end{figure*}

\begin{abstract}
Recent advances in single image-to-3D generation have enabled high-quality asset synthesis, yet extending these capabilities to indoor scene generation remains challenging. Existing methods focus on asset-level generation while neglecting the structural layout, which is essential for downstream applications and serves as the spatial anchor for grounding assets. However, a single image with a limited field of view lacks the spatial coverage to recover a coherent global layout. To this end, we use a 360° image represented in equirectangular projection (ERP) and propose \textbf{InSpace}, a structure-aware framework for 3D indoor scene generation. InSpace comprises three stages: (1) estimating partial scene geometry as spatial priors, (2) generating coarse scene structure with view-selective cross-attention, and (3) producing detailed layout and asset geometry with textures through a global-local hybrid attention, using flow matching. We also propose ERP-FRONT, a paired ERP-Image-to-3D indoor scene dataset based on 3D-FRONT. Experiments show that InSpace generates complete 3D indoor scenes with structural layout, along with separate textured assets from a single ERP image, achieving strong performance across 3D and 2D metrics.

\end{abstract}

\begin{figure}[t!]
\centering
    \includegraphics[width=\textwidth]{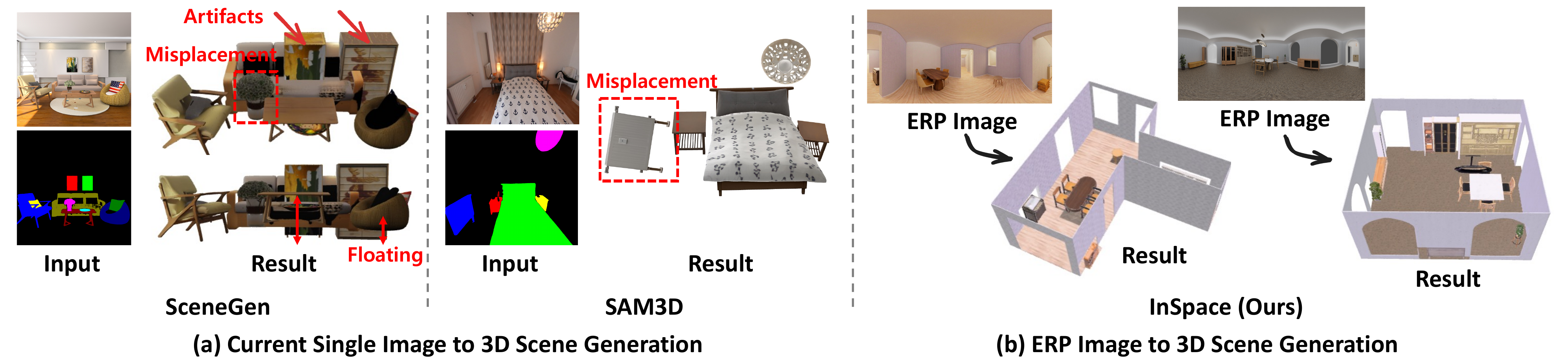}
   \caption{(a) Existing single-image methods generate individual assets without structural layout, causing floating, misplacement, and artifacts. (b) InSpace uses an ERP image to generate complete indoor scenes with structural layout and well-grounded assets.}
\label{fig:2_introduction}
\end{figure}

\section{Introduction}
\label{sec:intro}

Recent advances in generative modeling have dramatically accelerated 3D content creation from visual inputs. In particular, single-image-to-3D generation models \cite{liu2023zero, hong2024lrm, xu2024instantmesh, zhang2024clay, li2025step1x, zhao2023michelangelo, craftsman, lai2025unleashing, li2025triposg, zhao2025hunyuan3d, hunyuan3d2025hunyuan3d, xiang2025structured, ye2025hi3dgen, xiang2025native, wu2026direct3d, chen2026sam} have achieved remarkable quality and geometric fidelity, enabling rapid synthesis of an object from a single image. This success has naturally motivated extending these capabilities beyond isolated objects to entire 3D scene synthesis. Among these, indoor scene generation \cite{ling2025scenethesislanguagevisionagentic, gu2025artiscene, yang2026sceneweaver, yao2025cast, meng2026scenegen} has attracted growing attention due to its broad practical impact, from synthetic training environments for robotics and simulation, to digital twin creation for real estate and spatial planning, to enabling embodied agents to perceive and navigate 3D environments from a single observation.

However, existing methods focus on asset-level generation, producing individual assets and estimating their placement, while neglecting the structural layout of the indoor scene, including floors, walls, and built-in structures.
This structural layout is a fundamental component of indoor scene generation.  First, it reveals the overall shape and extent of the space, which is essential for any downstream task that requires scene-level spatial understanding. Second, it serves as a spatial anchor that constrains every object in the scene. Furniture rests on the floor, shelves lean against walls, and the structural geometry determines permissible placements. Without it, generated assets often exhibit spatial misalignment \cite{meng2026scenegen, huang2025midi}, floating above the ground or interpenetrating walls (Fig. \ref{fig:2_introduction}(a)), necessitating post-hoc physics-aware optimization \cite{ling2025scenethesislanguagevisionagentic, yao2025cast, yang2026sceneweaver}.

To address this limitation, we reformulate the problem by expanding the input modality from a single image to a 360° image represented in equirectangular projection (ERP). 
A single image provides only a limited field of view, making full structural layout estimation fundamentally ill-posed.
In contrast, an ERP image captures the complete 360° field of view of an indoor environment \cite{ai2022deep, ai2025survey}, offering comprehensive information about the scene geometry. Leveraging this richer spatial context, we introduce \textbf{InSpace}, a structure-aware framework for 3D indoor scenes, including both structural layout and individual assets from a single ERP image (Fig. \ref{fig:2_introduction} (b)).

InSpace comprises three cascaded stages. 
First, given a single ERP image, we estimate a depth map~\cite{li2025da2depthdirection} and lift it to 3D via equirectangular back-projection, producing an initial point cloud. This is then normalized into $[-0.5, 0.5]^3$, termed the Partial Scene Geometry (PSG), along with a calibrated camera center. Both serve as geometric priors for spatially grounding the subsequent generation stages.
Second, we convert the ERP image into six cubemap face images and use them as conditioning inputs to a flow-matching model~\cite{xiang2025native} that generates a coarse 3D voxel representation of the scene, termed the Coarse Scene Geometry (CSG). Unlike conventional 3D generation models, where every voxel latent is conditioned globally by a single image feature, our model introduces \textbf{view-selective cross-attention}, where each voxel in the latent is conditioned only on the cubemap faces visible from its 3D position based on the calibrated camera center. This spatially grounded conditioning enables the model to coherently compose the scene from all six cubemap views with fine-grained spatial control. From the resulting CSG, a lightweight 3D bounding box detector built on a 3D U-Net ~\cite{yin2021center, liu2022petr} predicts 3D oriented bounding boxes (3D OBBs) for each asset. 
Third, conditioned on the predicted 3D bounding boxes, we generate detailed geometry and texture for both the structural layout and each individual asset. This is achieved through a global-local hybrid attention mechanism. Global self-attention enables all scene components to interact for spatial coherence, while \textbf{asset-selective cross-attention} allows each component to attend only to its corresponding image region for fine-grained detail recovery. The resulting latents are finally decoded into a textured 3D mesh of the complete indoor scene.

To train and evaluate InSpace, we require paired ERP images and corresponding 3D indoor scene meshes. However, such data is unavailable in existing benchmarks. While several real-world panoramic datasets~\cite{chang2017matterport3d, zheng2020structured3d, straub2019replica, dong2024panocontext} provide ERP images of indoor scenes, their reconstructed meshes are often incomplete or lack the structural fidelity necessary. To address this gap, we construct \textbf{ERP-FRONT}, a synthetic ERP-Image-to-3D indoor scene dataset based on 3D-FRONT~\cite{fu20213d-3d-front}. Each indoor scene is defined at the room level, covering a wide range of room sizes, and paired with ERP observations rendered from within the scene. The resulting dataset contains 26.5K training and 2.5K test ERP-image-mesh pairs. In our experiments, InSpace shows strong performance on both 3D geometric and 2D rendering-based metrics. Moreover, as shown in the supplementary material, InSpace generalizes well to realistic panoramic datasets such as ReplicaPano~\cite{dong2024panocontext}, producing spatially coherent 3D indoor environments.

\section{Related Work}

\subsubsection{3D Asset Generation}
Following the success of latent diffusion in 2D\cite{rombach2022high, labs2025flux}, recent 3D asset generation methods adopt VAEs \cite{kingma2013auto, razavi2019generating} to encode 3D geometry into latent spaces, where diffusion \cite{peebles2023scalable} or flow-based models \cite{lipman2022flow} learn to generate. Depending on the VAE design, existing approaches fall into two categories: vector-set (VecSet) and sparse voxel representations.
\textbf{VecSet representations}, introduced in 3DShape2VecSet \cite{zhang20233dshape2vecset}, encode 3D shapes as unordered latent token sets derived from surface point sampling. A set of learnable queries \cite{carion2020end, jaegle2021perceiver} interacts with point embeddings via cross-attention to form an orthogonal basis of the shape latent space, capturing relative geometric relationships rather than absolute spatial coordinates. This design is efficient and transformer-friendly, leading to wide adoption in large-scale generation models \cite{zhang2024clay, li2025step1x, zhao2023michelangelo, craftsman, lai2025unleashing, li2025triposg, zhao2025hunyuan3d, hunyuan3d2025hunyuan3d}. However, since tokens are not anchored to fixed spatial locations, VecSet lacks explicit spatial locality, making position-aware control and structured test-time scaling challenging.
By contrast, \textbf{sparse voxel representations} encode 3D shapes as structured grids with explicit spatial coordinates. Introduced in XCube \cite{ren2024xcube}, sparse voxel VAEs \cite{he2025sparseflex, li2026sparc3d} compress 3D occupancy or feature grids via 3D convolutional encoders, maintaining spatially organized latent tensors aligned with 3D coordinates rather than learning an abstract token basis. This structured design enables locality for position-aware conditioning and local editing, and has been adopted in high-fidelity models \cite{xiang2025structured, ye2025hi3dgen, xiang2025native, wu2026direct3d, chen2026sam, li2026pixal3d}. However, sparse voxel methods are more computationally expensive and tightly coupled to grid resolution, limiting scalability compared to VecSet-based approaches. Recent work \cite{chen2025ultra3d, lai2026lattice, jia2025ultrashape10highfidelity3d} explores hybrid representations combining structured voxel anchors with VecSet-based latents to balance efficiency and spatial locality.

\subsubsection{3D Scene Generation}
Early methods \cite{gao2024diffcad, gumeli2022roca, kuo2021patch2cad, kuo2020mask2cad} relied on retrieving and aligning existing 3D models to the input image, but were limited by dataset diversity and sparse geometric cues. The advent of image-to-3D models enabled a new paradigm of generating objects individually and assembling them via language/vision-guided spatial reasoning \cite{ling2025scenethesislanguagevisionagentic, gu2025artiscene, yang2026sceneweaver}, and physics-based optimization \cite{ling2025scenethesislanguagevisionagentic, yao2025cast, yang2026sceneweaver} proposed to improve placement. However, these multi-stage pipelines are susceptible to error accumulation and require costly optimization for object placement. Recent simultaneous multi-instance generation methods \cite{lin2025partcrafterstructured3dmesh, huang2025midi, meng2026scenegen} improve inter-object coherence through cross-instance interaction, yet remain focused on asset-level synthesis without modeling structural layouts (e.g., floors, walls, ceilings), causing spatial misalignment such as floating or interpenetrating objects. HiScene \cite{dong2025hiscene} addresses this by converting inputs into isometric views and generating compositional structures to ground objects on architectural surfaces, but the resulting layouts remain approximate and do not capture precise indoor geometry. Most related to our work, PanoContext-Former \cite{dong2024panocontext} is the first attempt at ERP-Image-to-indoor-scene generation, but produces inaccurate geometry, lacks texture support, and its code is unavailable.

\section{Preliminary}
\label{sec:preliminary}

Our method requires explicit spatial control over 3D voxel latents using ERP image features, necessitating a spatially structured latent space rather than an unordered VecSet representation. We therefore build upon TRELLIS.2~\cite{xiang2025native}, which adopts the O-Voxel formulation to encode geometry and appearance within a compact, sparse voxel latent space.

% \paragraph{\textbf{O-Voxel Representation.}}
\subsubsection{O-Voxel Representation.}
O-Voxel encodes both geometry and appearance as a set of feature tuples associated with active voxels on a regular $N^3$ grid:
\begin{equation}
    \mathbf{f} = \{(\mathbf{f}^{\text{shape}}_i, \mathbf{f}^{\text{mat}}_i, \mathbf{p}_i)\}_{i=1}^{L},
    \label{eq:1}
\end{equation}
where $\mathbf{f}^{\text{shape}}_i$ encodes local geometry via a Flexible Dual Grid formulation, $\mathbf{f}^{\text{mat}}_i$ encodes PBR material properties with structural layout and textured assets (\ie, base color, metallic, roughness, and opacity), $\mathbf{p}_i \in \{0,\dots,N\!-\!1\}^3$ is the voxel coordinate, and $L \ll N^3$ is the number of active (surface-intersecting) voxels.

\subsubsection{Sparse Compression VAE.}
To obtain a compact latent space, the framework trains two decoupled Sparse Compression VAEs (SC-VAEs) $\{\mathcal{E}_{\text{shape}}, \mathcal{D}_{\text{shape}}\}$ and $\{\mathcal{E}_{\text{mat}},\allowbreak \mathcal{D}_{\text{mat}}\}$ for shape and material, respectively. Both are fully sparse-convolutional networks with a spatial downsampling factor $s$, compressing each active voxel's features into a latent vector of $c_{sc}=32$ channels while preserving the sparse coordinate structure at the downsampled resolution $\frac{N}{s}$.
% do not forget to mention it in the implementation section! -> $16\times$ %

\subsubsection{Three-Stage Generation Pipeline.}
% Given a conditioning image $\mathbf{I}$, the framework generates 3D assets through three sequential stages, each employing a DiT-based flow matching model that denoises Gaussian noise $\boldsymbol{\epsilon}$ over timestep $t$:
In TRELLIS.2, given a conditioning image $\mathbf{I}$, 3D assets are generated through three sequential stages, each using a DiT-based flow matching model \cite{peebles2023scalable, lipman2022flow} that denoises Gaussian noise $\boldsymbol{\epsilon}$ over timestep $t$:
\begin{align}
    &\textit{Sparse Structure:} ~~ \mathcal{F}_{\text{ss}}\!: (\boldsymbol{\epsilon}, t, \mathbf{I}) \rightarrow \mathbf{z}_{\text{ss}}; ~~ \mathbf{p} = \mathcal{D}_{\text{ss}}(\mathbf{z}_{\text{ss}}), \label{eq:2} \\
    &\textit{Geometry:} ~~ \mathcal{F}_{\text{shape}}\!: (\boldsymbol{\epsilon}_{\text{shape}}, t, \mathbf{I}) \rightarrow \mathbf{z}_{\text{shape}}; ~~ \mathbf{f}^{\text{shape}} = \mathcal{D}_{\text{shape}}(\mathbf{z}_{\text{shape}}), \label{eq:3} \\
    &\textit{Material:} ~~ \mathcal{F}_{\text{mat}}\!: (\boldsymbol{\epsilon}_{\text{mat}}, t, \mathbf{I}, \mathbf{z}_{\text{shape}}) \rightarrow \mathbf{z}_{\text{mat}}; ~~ \mathbf{f}^{\text{mat}} = \mathcal{D}_{\text{mat}}(\mathbf{z}_{\text{mat}}). \label{eq:4}
\end{align}
In the \textit{sparse structure generation} stage (Eq.~\ref{eq:2}), DiT $\mathcal{F}_{\text{ss}}$ generates a structure latent $\mathbf{z}_{\text{ss}} \in \mathbb{R}^{c_0 \times \frac{N_0}{s_0}^3}$ from noise $\boldsymbol{\epsilon}$, where $c_0$ is the latent channel dimension, $N_0$ is the grid resolution, and $s_0$ is the downsampling factor (\eg, $c_0\!=\!8$, $N_0\!=\!64$, $s_0\!=\!4$). The decoder $\mathcal{D}_{\text{ss}}$ maps $\mathbf{z}_{\text{ss}}$ into a coarse occupancy grid of resolution $N_0^3$ to determine the set of active voxel coordinates~$\mathbf{p}$.
In the \textit{geometry generation} stage (Eq.~\ref{eq:3}), structured noise $\boldsymbol{\epsilon}_{\text{shape}} = \{(\boldsymbol{\epsilon}_i, \mathbf{p}_i)\}_{i=1}^{L}$ is formed as a SparseTensor by assigning random Gaussian vectors $\boldsymbol{\epsilon}_i \in \mathbb{R}^{c_{sc}}$ to each active coordinate, and a sparse DiT $\mathcal{F}_{\text{shape}}$ denoises it into the geometry latent $\mathbf{z}_{\text{shape}}$, where voxel coordinates $\{p_i\}_{i=1}^{L}$ are injected via positional embeddings.
The \textit{material generation} stage (Eq.~\ref{eq:4}) follows the same sparse formulation but additionally conditions on $\mathbf{z}_{\text{shape}}$ via channel-wise concatenation.
The decoded O-Voxel is then instantly converted into a textured mesh.

\begin{figure}[t!]
\centering
    \includegraphics[width=1.0\textwidth]{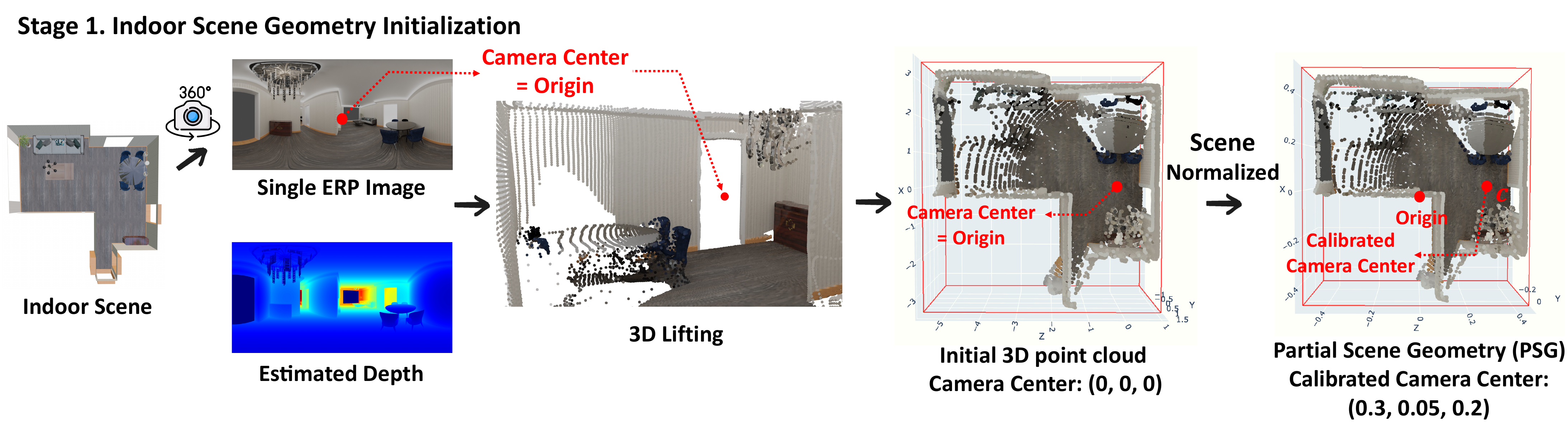}
   \caption{\textbf{Indoor Scene Geometry Initialization.} From a single ERP image and its estimated depth, we lift the scene to 3D via equirectangular back-projection, producing an initial point cloud with the camera center at the origin $(0,0,0)$. The point cloud is then normalized into the canonical space $[-0.5, 0.5]^3$, yielding the Partial Scene Geometry (PSG) and shifting the camera center to a calibrated position $\mathbf{c}$.}
\label{fig:3}
\end{figure}

\section{Method}
\label{sec:blind}
Given a single ERP image, InSpace generates a complete 3D indoor scene comprising both the scene layout (floors and walls) and individual assets with detailed geometry and texture. To enable spatial control over voxel latents conditioned on ERP features, we adopt a sparse voxel-based framework \cite{xiang2025native} and further decompose the scene into layout and asset-level components inspired by part-aware generation \cite{yang2025omnipart, ding2025fullpartgenerating3dresolution}.
Our framework consists of three stages. First, we estimate the scene layout by lifting the ERP depth map to a 3D point cloud and calibrating the camera center within a normalized voxel space (Sec. \ref{sec:method_step1}). Second, we generate a coarse scene structure via flow-based denoising in a dense voxel latent space, conditioned on cubemap features through view-selective cross-attention, and detect 3D oriented bounding boxes (OBBs) of assets from the generated structure (Sec. \ref{sec:method_step2}). Third, we produce detailed geometry and texture through a hybrid attention mechanism combining global self-attention with OBB-projected asset-selective cross-attention. Each asset attends to its relevant image region while preserving scene-level coherence (Sec. \ref{sec:method_step3}).

\subsection{Stage 1. Indoor Scene Geometry Initialization}
\label{sec:method_step1}
We begin by constructing the geometric foundation of the scene from a single ERP image. Using a pretrained depth estimator \cite{li2025da2depthdirection}, we obtain a dense depth map and lift it to 3D via equirectangular projection, producing an initial point cloud. Since the ERP image is captured from a single viewpoint, the camera center (the position from which the panorama was rendered) lies at the origin $(0,0,0)$.
To prepare for voxel-based generation, we normalize the point cloud into the canonical space $[-0.5, 0.5]^3$ via translation and uniform scaling. This normalization shifts the camera center from the origin to a new position $\mathbf{c}$ within the canonical space (see Fig. \ref{fig:3}). We refer to the result as the \textit{Partial Scene Geometry (PSG)}, as it captures only the surfaces visible from the single viewpoint. The calibrated camera center $\mathbf{c}$ and PSG are passed to subsequent stages as geometric priors for spatially grounding the generation process.

\begin{figure}[t!]
    \centering
    \includegraphics[width=\linewidth]{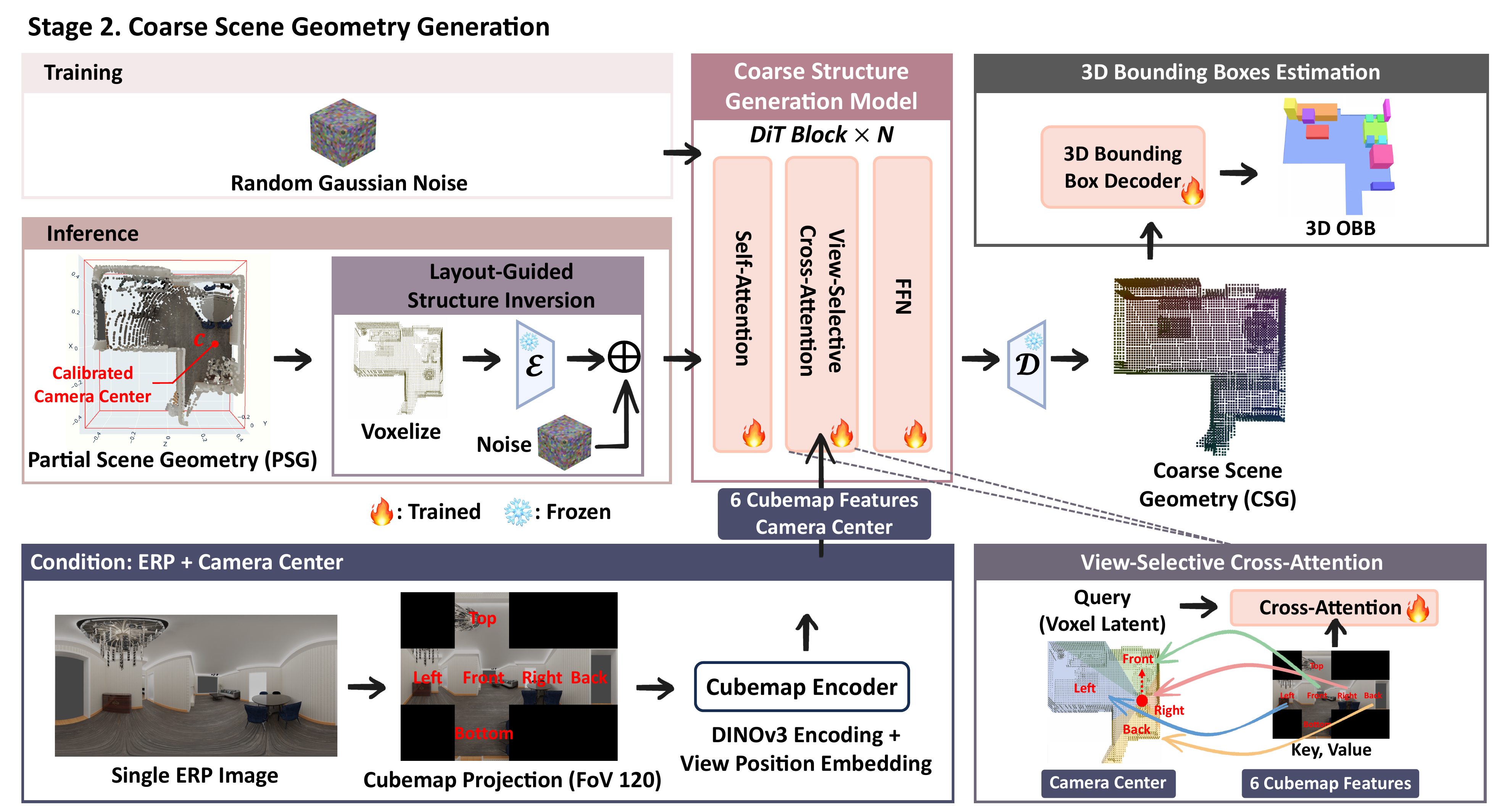}
    \caption{\textbf{Coarse Scene Geometry Generation.} Six cubemap faces (FoV $120°$) are extracted from the ERP image and encoded with view position embeddings. The model denoises voxel latents via DiT blocks with view-selective cross-attention, where each voxel attends only to its visible faces based on camera center $\mathbf{c}$. At inference, Layout-Guided Structure Inversion optionally initializes the latent from the PSG. The decoded coarse geometry (CSG) is fed to a 3D bounding box detector to predict OBBs for assets.}
    \label{fig:4}
\end{figure}

\subsection{Stage 2. Coarse Scene Geometry Generation}
\label{sec:method_step2}
In this stage, we generate the coarse structure of the indoor scene as a dense voxel latent $\mathbf{z}_{\text{ss}} \in \mathbb{R}^{c_0 \times (N_0/s_0)^3}$ (Eq. \ref{eq:2}) using a flow matching model \cite{xiang2025native}. The model consists of stacked DiT blocks, each applying self-attention among voxel latents, followed by view-selective cross-attention with the six cubemap features, and a feed-forward network (Fig. \ref{fig:4}). It is trained with rectified flow matching (Sec.~\ref{sec:method_step4}) to predict the velocity field along the path connecting Gaussian noise and the target voxel latent.

\subsubsection{Cubemap Encoding.}
To condition the model on the input ERP image, we extract six cubemap faces along the six orthogonal viewing directions, with field of view (FoV) set to $120°$ to introduce overlap between adjacent faces and prevent objects near boundaries from being clipped. Each face is independently encoded by a DINOv3 \cite{simeoni2025dinov3} image encoder, producing $T_f$ tokens of dimension $D_c$ per face, where $f$ indexes the six cubemap faces. To distinguish face directions, we add a learnable view position embedding $\mathbf{e}_v \in \mathbb{R}^{6 \times D_c}$ to each face's features. The six encoded faces are concatenated along the token dimension, yielding a unified condition tensor $\mathbf{F} \in \mathbb{R}^{B \times T \times D_c}$, where $B$ is the batch size and $T = 6 T_f$.

\subsubsection{View-Selective Cross-Attention.}
In existing 3D asset generation~\cite{xiang2025native, hunyuan3d2025hunyuan3d}, a single condition image influences all voxels in the latent uniformly during denoising. In our panoramic setting, however, the six cubemap faces each correspond to a distinct spatial region, and voxels in different regions of $\mathbf{z}_{\text{ss}}$ (Eq. \ref{eq:2}) should be conditioned primarily on their corresponding faces rather than all condition tokens. To this end, we construct a view-selective (VS) cross-attention mask $\mathbf{M}_{\text{vs}}$ (Eq.~\ref{eq:6_vs_mask}) that establishes this spatially grounded correspondence.

\begin{figure}[t!]
\centering
    \includegraphics[width=\textwidth]{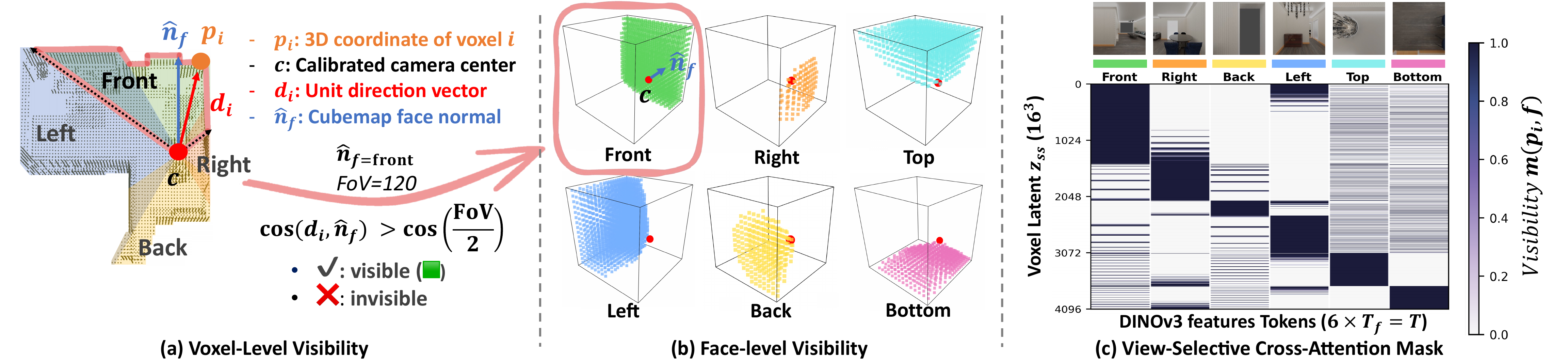}
   \caption{\textbf{View-Selective Cross-Attention.} (a) For each voxel at position $p_i$, we compute the cosine similarity between the direction $d_i$ from camera center c and each cubemap face normal $\hat{n}_f$ to determine visibility. (b) Voxels are partitioned by their visible faces, shown in corresponding colors. (c) The resulting mask $M_{vs}$ assigns each voxel high attention weights (dark) only to its visible cubemap tokens.}
\label{fig:5}
\end{figure}

For the $i$-th voxel at coordinate $\mathbf{p}_i \in \{0,\dots,N_0/s_0-1\}^3$ (Eq.~\ref{eq:1}), we compute the unit direction $\mathbf{d}_i = (\mathbf{p}_i - \mathbf{c}) / \|\mathbf{p}_i - \mathbf{c}\|$ from the calibrated camera center $\mathbf{c}$ toward the voxel, and measure how well this direction aligns with each cubemap face $f$, 
whose outward normal is $\hat{\mathbf{n}}_f$  (e.g., $\hat{\mathbf{n}}_{\text{front}} = (0,1,0)$).
The cosine similarity $\mathbf{d}_i \cdot \hat{\mathbf{n}}_f$ is compared against $\cos(\text{FoV}/2)$, the angular boundary of each face's visible cone. To produce a smooth, differentiable mask rather than a hard cutoff at face boundaries, we pass this comparison through a scaled sigmoid:
\begin{align}
    m(\mathbf{p}_i, f; \mathbf{c}) = \sigma\!\left(\alpha\!\left(\mathbf{d}_i \cdot \hat{\mathbf{n}}_f - \cos\!\left(\tfrac{\text{FoV}}{2}\right)\right)\right) \in (0, 1], \label{eq:5_visibility}
\end{align}
where $\alpha$ is a fixed scaling factor that yields a near-binary transition. The output $m$ represents visibility (see Fig. \ref{fig:5}(a)), where values near 1 indicate the voxel falls well within the face's FoV and values near 0 indicate it lies outside. This soft assignment effectively partitions $\mathbf{z}_{\text{ss}}$ into spatially coherent groups, each primarily conditioned on its corresponding cubemap face, while allowing smooth transitions at face boundaries due to the overlapping FoV.
The face-level visibility is broadcast to all $T_f$ tokens within each face to form the full mask (Fig. \ref{fig:5}(b,c)):
\begin{align}
    \mathbf{M}_{\text{vs}}[i, j] &= m(\mathbf{p}_i,\, f(j);\, \mathbf{c}), \quad \mathbf{M}_{\text{vs}} \in \mathbb{R}^{(N_0/s_0)^3 \times T}, \label{eq:6_vs_mask}
\end{align}
where $i$ indexes voxel latents, $j$ indexes condition tokens, and $f(j)$ maps the $j$-th token to its cubemap face. Each row encodes a single voxel's visibility across all $T$ tokens, and each column reflects the visibility of its corresponding cubemap face. This mask is applied as an additive bias to the cross-attention logits:
\begin{align}
    \text{VS-CrossAttn}(\mathbf{Q}, \mathbf{K}, \mathbf{V}; \mathbf{M}_{\text{vs}}) = \text{softmax}\!\left(\frac{\mathbf{Q}\mathbf{K}^\top}{\sqrt{d_h}} + \log \mathbf{M}_{\text{vs}}\right)\mathbf{V}. \label{eq:7_vsattn}
\end{align}
where $d_h$ is the attention head dimension, $\mathbf{Q} \in \mathbb{R}^{(N_0/s_0)^3 \times d_h}$ are voxel queries projected from $\mathbf{z}_{\text{ss}}$ (Eq. \ref{eq:2}), and $\mathbf{K}, \mathbf{V} \in \mathbb{R}^{T \times d_h}$ are keys and values projected from $\mathbf{F}$. Since $m \in (0, 1]$, the $\log\mathbf{M}_{\text{vs}}$ term acts as a soft penalty, where near-zero visibility yields large negative values that suppress attention weights, while full visibility contributes zero and leaves them unchanged.

\begin{figure}[t!]
\centering
    \includegraphics[width=1.0\textwidth]{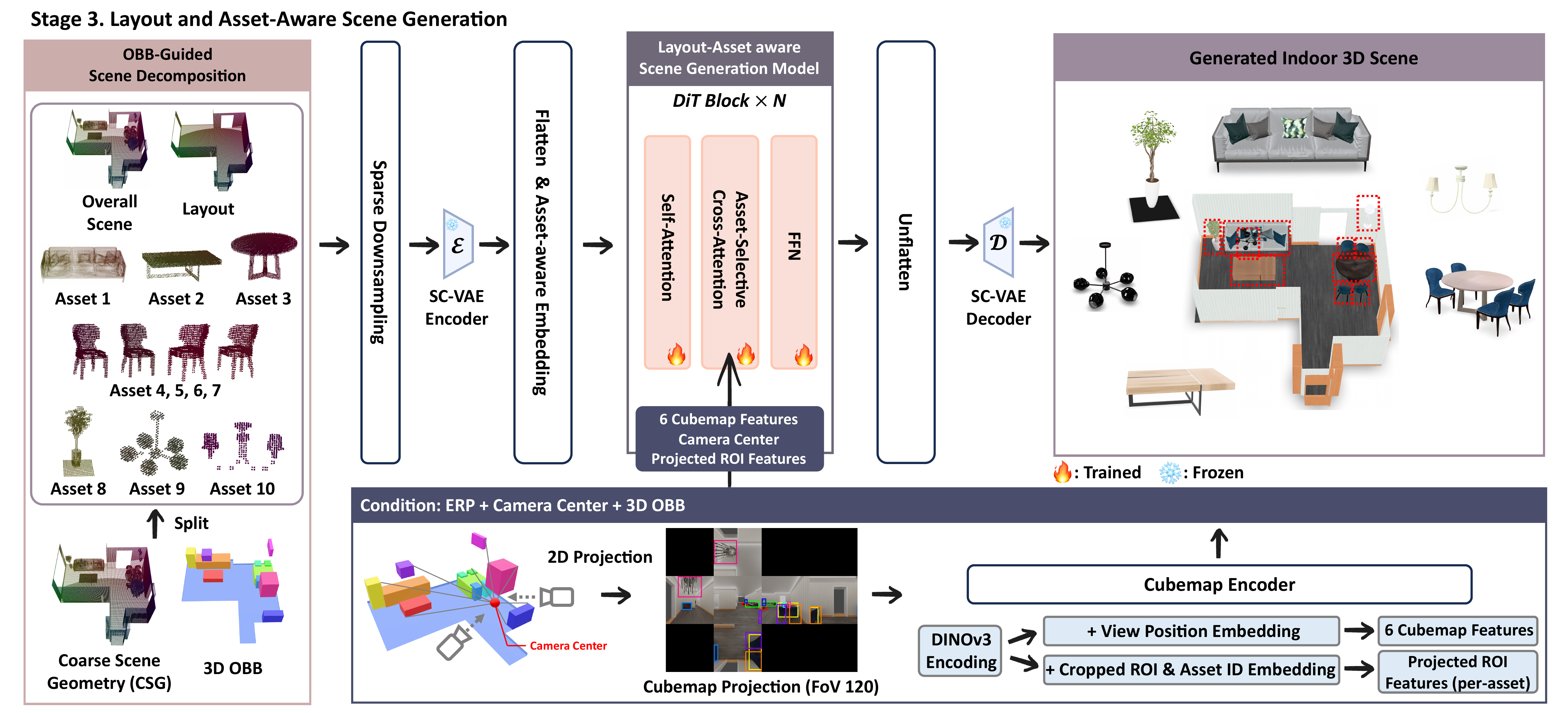}
   \caption{\textbf{Layout and Asset-Aware Scene Generation.} The CSG is decomposed into layout and assets via detected OBBs, encoded with part embeddings, and concatenated into a unified token sequence. DiT blocks apply global self-attention, followed by asset-selective cross-attention where layout tokens use the view-selective mask and each asset attends to its projected ROI features. The output is decoded into a textured 3D scene.}
\label{fig:6}
\end{figure}

\subsubsection{Layout-Guided Structure Inversion.}
% At inference, we leverage the PSG from Sec.~\ref{sec:method_step1} as a spatial prior. Rather than starting from pure Gaussian noise, we voxelize the PSG and encode it via $\mathcal{E}_{\text{ss}}$ into $\mathbf{z}_{\text{psg}}$, then construct the noised initial latent $\mathbf{x}_{t_0} = (1 - t_0)\,\mathbf{z}_{\text{psg}} + t_0\,\boldsymbol{\epsilon}$, $\boldsymbol{\epsilon} \sim \mathcal{N}(\mathbf{0}, \mathbf{I})$, and denoise from $t_0$ to $0$. Inspired by SDEdit~\cite{meng2021sdedit}, a lower $t_0$ preserves more of the PSG layout while a higher $t_0$ allows greater model freedom. During training, the model learns from pure noise ($t_0\!=\!1$) to maximize generalization. 
At inference, we leverage the PSG from Sec.~\ref{sec:method_step1} as a spatial prior. This is inspired by SDEdit~\cite{meng2021sdedit} and other diffusion-based image editing methods \cite{lugmayr2022repaint, couairon2023diffedit, shi2024dragdiffusion, koo2024flexiedit, koo2025flowdrag, yoon2024dni}, which add noise to a guidance input and denoise from that intermediate noise level rather than from pure noise to keep the output faithful to the input. We voxelize the PSG and encode it via $\mathcal{E}_{\text{ss}}$ into $\mathbf{z}_{\text{psg}}$, then construct the noised initial latent $\mathbf{x}_{t_0} = (1 - t_0)\,\mathbf{z}_{\text{psg}} + t_0\,\boldsymbol{\epsilon}$, $\boldsymbol{\epsilon} \sim \mathcal{N}(\mathbf{0}, \mathbf{I})$, and denoise from $t_0$ to $0$. A lower $t_0$ preserves more of the PSG layout, while a higher $t_0$ allows greater model freedom. This requires no change to training, as the model is optimized with the standard flow-matching objective over $t \sim \mathcal{U}(0,1)$ and thus learns to denoise the target latent from any noise level, allowing inference to begin from the noised PSG at $t_0 < 1$ instead of pure noise ($t_0 = 1$). Since the PSG already encodes the geometry recovered from the input ERP, initializing from this information-rich state rather than from pure noise provides a strong spatial anchor and yields more accurate scene structure.

\subsubsection{3D Bounding Box Detection.}

After denoising by the stacked DiT blocks, we obtain the generated structure latent $\mathbf{z}_{\text{ss}}$ and decode it via $\mathcal{D}_{\text{ss}}$ (Eq.~\ref{eq:2}) into the \textit{Coarse Scene Geometry (CSG)}, a binary occupancy grid at resolution $N_0^3$. A 3D U-Net–based detector~\cite{yin2021center, liu2022petr} then predicts $K$ OBBs $\{\mathbf{o}_k\}_{k=1}^{K}$ in a single pass, each parameterized by its center $\mathbf{o}_k^c \in [-0.5, 0.5]^3$, extents $\mathbf{o}_k^s \in [0, 0.5]^3$, and yaw angle $\theta_k \in [-\pi, \pi]$. Architectural details are in Sec.~\ref{sec:supple_2_bbox} of the supplementary.

\subsection{Stage 3. Layout and Asset-Aware Scene Generation}
\label{sec:method_step3}

The final stage generates detailed geometry and texture for the complete indoor scene (Eqs.~\ref{eq:3}--\ref{eq:4}). Inspired by part-aware 3D generation~\cite{yang2025omnipart}, we decompose the scene into layout and asset-level components and process them jointly through a hybrid attention mechanism. Shape and texture share an identical model architecture, where the texture model additionally conditions on the generated shape latent $\mathbf{z}_{\text{shape}}$ via concatenation (Eq.~\ref{eq:4}).

\subsubsection{OBB-Guided Scene Decomposition.}
Using the detected OBBs from Sec.~\ref{sec:method_step2}, we decompose the CSG into individual components (Fig.~\ref{fig:6}, left): the overall scene (all active voxels), the scene layout (voxels outside any OBB, representing floors and walls), and $K$ individual assets (voxels enclosed within each $\mathbf{o}_k$). Each component is encoded by the SC-VAE encoder $\mathcal{E}$ (Sec.~\ref{sec:preliminary}), flattened into a token sequence, and summed with a learnable part positional embedding $\mathbf{e}_p \in \mathbb{R}^{D}$ (DiT hidden dim) encoding its component identity. All sequences are concatenated into a unified token sequence and processed by stacked DiT blocks, where each block applies global self-attention followed by asset-selective cross-attention.
% part positional embedding -> it looks fine, but need to check

\subsubsection{Global Scene Self-Attention.}
Within each DiT block, self-attention is first applied across all components in the unified token sequence, enabling the model to learn inter-object spatial relationships and physical placement constraints~\cite{yang2025omnipart}.

\subsubsection{Asset-Selective Cross-Attention.}
While self-attention operates globally, cross-attention is scoped per component, directing each to its relevant image region. Since the view-selective mask (Eq.~\ref{eq:5_visibility}) depends only on the calibrated camera center $\mathbf{c}$ and voxel coordinates, it generalizes to any voxel latent beyond $\mathbf{z}_{\text{ss}}$. We leverage this to construct a unified cross-attention mask $\mathbf{M}_{\text{as}} \in \mathbb{R}^{N_{\text{total}} \times T}$, where $N_{\text{total}}$ is the total number of voxels across all components and $T$ is the number of cubemap condition tokens (Sec.~\ref{sec:method_step2}), combining two masking strategies:
\begin{align}
    \mathbf{M}_{\text{as}}[i, j] = 
    \begin{cases} 
        \mathbf{M}_{\text{vs}}[i, j] \in \mathbb{R}^{(N_{\text{ov}} + N_{\text{la}}) \times T} & \text{if voxel } i \in \text{overall or layout}, \\
        \mathbf{M}_{\text{as}}^{(k)}[i, j] \in \{0,1\}^{N_k \times T} & \text{if voxel } i \in \text{asset } k,
    \end{cases}
    \label{eq:8_as_mask}
\end{align}
where $N_{\text{ov}}$, $N_{\text{la}}$, and $N_k$ denote the number of voxels in the overall scene, layout, and asset $k$ respectively. The full mask $\mathbf{M}_{\text{as}}$ is formed by stacking $\mathbf{M}_{\text{vs}}$ for the overall and layout components with $\mathbf{M}_{\text{as}}^{(k)}$ for each asset along the voxel dimension, yielding $N_{\text{total}} = N_{\text{ov}} + N_{\text{la}} + \sum_{k=1}^{K} N_k$.
For the overall scene and layout components, $\mathbf{M}_{\text{vs}}[i,j]$ reuses the view-selective visibility $m(\mathbf{p}_i, f(j); \mathbf{c})$ (Eq.~\ref{eq:5_visibility}) applied to the voxel coordinates of $\mathbf{z}_{\text{shape}}$ and $\mathbf{z}_{\text{mat}}$ (Eqs.~\ref{eq:3}--\ref{eq:4}), so that each voxel attends to the cubemap faces visible from its 3D position.
For each asset $k$, $\mathbf{M}_{\text{as}}^{(k)}$ is defined via $m_{\text{proj}}(\mathbf{o}_k, j; \mathbf{c}) \in \{0,1\}$, obtained by projecting the eight corners of $\mathbf{o}_k$ onto the six cubemap planes using $\mathbf{c}$ and the FoV geometry (Fig.~\ref{fig:6}, bottom). The resulting 2D bounding rectangles crop the corresponding regions from $\mathbf{F}$, setting $m_{\text{proj}}=1$ for tokens within the region and $0$ otherwise. An asset identity embedding is added to the cropped tokens, so that when $M^{k}_{as}$ is stacked into the unified $M_{as}$, the model can identify the asset associated with each token. Since 3D OBB projection depends only on $\mathbf{o}_k$, this mask is shared across all voxels within asset $k$. The mask $M_{as}$ is added to the attention logits:
\begin{align}
    \text{AS-CrossAttn}(\mathbf{Q}, \mathbf{K}, \mathbf{V}; \mathbf{M}_{\text{as}}) = \text{softmax}\!\left(\frac{\mathbf{Q}\mathbf{K}^\top}{\sqrt{d_h}} + \log \mathbf{M}_{\text{as}}\right)\mathbf{V} \label{eq:9_asattn}.
\end{align}
where $\mathbf{Q} \in \mathbb{R}^{N_{\text{total}} \times d_h}$ are voxel queries from the unified token sequence, and $\mathbf{K}, \mathbf{V} \in \mathbb{R}^{T \times d_h}$ are keys and values projected from $\mathbf{F}$. This enables layout components to attend to their geometrically visible cubemap faces via $\mathbf{M}_{\text{vs}}$, while each asset $k$ attends exclusively to its OBB-projected image region via $\mathbf{M}_{\text{as}}^{(k)}$, recovering fine-grained appearance details. The outputs are unflattened and decoded by $\mathcal{D}$ to produce the final textured 3D mesh.

\subsection{Training Objective}
\label{sec:method_step4}
All three models $\mathcal{F}_{\text{ss}}$, $\mathcal{F}_{\text{shape}}$, and $\mathcal{F}_{\text{mat}}$ (Eqs.~\ref{eq:2}--\ref{eq:4}) are trained with rectified flow matching \cite{liu2023flow}. Given a ground-truth latent $\mathbf{x}_0$ and noise $\boldsymbol{\epsilon} \sim \mathcal{N}(\mathbf{0}, \mathbf{I})$, we construct the interpolant $\mathbf{x}_t = (1-t)\,\mathbf{x}_0 + t\,\boldsymbol{\epsilon}$ at $t \sim \mathcal{U}(0,1)$ and train the model to predict the velocity field:
$\mathcal{L} = \mathbb{E}_{t,\, \mathbf{x}_0,\, \boldsymbol{\epsilon}}\!\left[\left\|\, \mathbf{v}_\theta\!\left(\mathbf{x}_t,\, t,\, \mathbf{F},\, \mathbf{M}\right) - \left(\boldsymbol{\epsilon} - \mathbf{x}_0\right)\right\|^2\right]$
where $\mathbf{F}$ denotes the cubemap condition features and $\mathbf{M}$ is the stage-specific attention mask: $\mathbf{M}_{\text{vs}}$ (Eq.~\ref{eq:6_vs_mask}) for $\mathcal{F}_{\text{ss}}$, and $\mathbf{M}_{\text{as}}$ (Eq.~\ref{eq:8_as_mask}) for $\mathcal{F}_{\text{shape}}$ and $\mathcal{F}_{\text{mat}}$. For material generation, $\mathbf{v}_\theta$ additionally conditions on the generated shape latent $\mathbf{z}_{\text{shape}}$ via concatenation (Eq.~\ref{eq:4}).

\begin{figure}[t!]
\centering
    \includegraphics[width=1.0\textwidth]{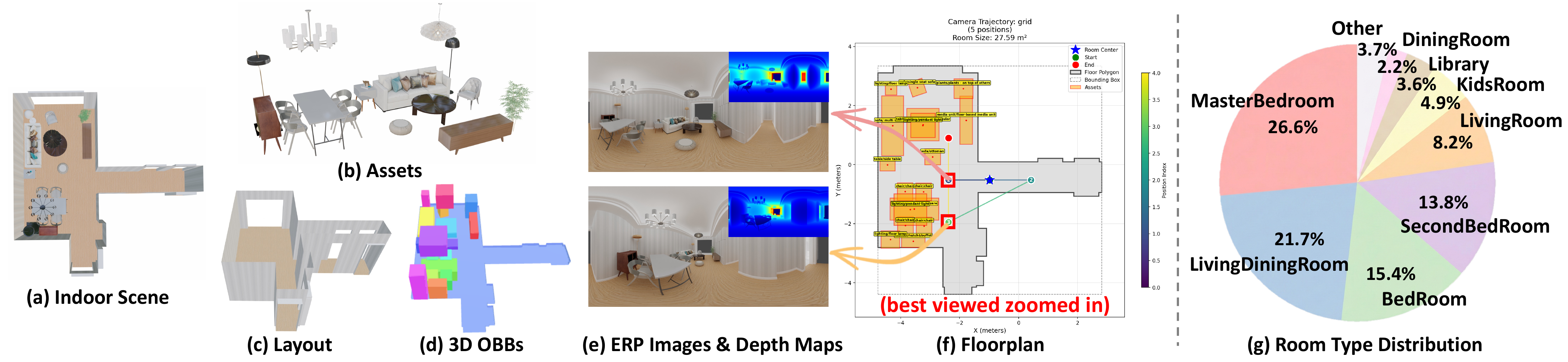}
   \caption{\textbf{ERP-FRONT Dataset.} Each room from 3D-FRONT is decomposed into (a) the complete indoor scene, (b) individual assets, (c) structural layout, and (d) 3D oriented bounding boxes. (e) ERP images with depth maps are rendered from valid camera positions in (f) the floorplan. (g) Room type distribution across the dataset.}
\label{fig:7}
\end{figure}

\section{ERP-FRONT Dataset}
\label{sec:dataset}
We construct ERP-FRONT, a paired ERP-Image-to-3D indoor scene dataset built on 3D-FRONT \cite{fu20213d-3d-front}, rendered using BlenderProc \cite{denninger2019blenderproc}. Each house is decomposed at the room level (one or more rooms per house), and for each room we extract the structural layout (floors and walls), individual assets with their 3D OBBs, and a floorplan encoding asset positions (Fig.\ref{fig:7}(a)-(f)).
To render ERP images, we search for valid camera positions on a uniform grid within each room, requiring a minimum clearance of 30\,cm from all assets to reduce occlusion. 
The resulting dataset comprises 5,962 houses and 13,292 rooms, totaling approximately 29K ERP-Image-mesh pairs. The average room area is 19.8\,m$^2$ with 8.4 assets per room. We split the data into 12,121 training rooms (26.5K ERP-image-mesh pairs) and 1,144 test rooms (2.5K pairs). Room type distribution is shown in Fig.~\ref{fig:7}(g). More details are provided in Sec.~\ref{sec:supple_1_erpfront} of the supplementary.

\section{Experiments}
\label{sec:experiments}

\subsection{Implementation Details}
\label{sec:impl_details}
Our framework is built upon TRELLIS.2~\cite{xiang2025native}, using all VAE encoders and decoders with pretrained weights without finetuning. For sparse structure generation (Eq.~\ref{eq:2}), we set $c_0=8$, $N_0=64$, $s_0=4$, yielding a latent resolution of $16^3$. For shape and texture generation (Eqs.~\ref{eq:3}--\ref{eq:4}), the SC-VAE uses $c_{\text{sc}}=32$ channels with downsampling factor $s=16$ on an $N=512$ grid, producing sparse latents at $32^3$ resolution. All stages use a frozen DINOv3 ViT-L/16~\cite{simeoni2025dinov3} encoder ($512 \times 512$ input, $T_f=1029$ tokens, $D_c=1024$).
The scaling factor in Eq. \ref{eq:5_visibility} is set to $\alpha$=50. 
Each generation model has 30 DiT blocks ($D=1536$, 12 heads, $d_h=128$) finetuned from pretrained weights with AdamW (lr $5 \times 10^{-5}$). $\mathcal{F}_{\text{ss}}$, $\mathcal{F}_{\text{shape}}$, and $\mathcal{F}_{\text{mat}}$ are trained for 120, 60, and 24 epochs respectively on 6$\times$ NVIDIA A100 GPUs with batch size 4 per GPU. The 3D bounding box detector is a lightweight 3D U-Net~\cite{yin2021center} (12.5M parameters) trained on the decoded CSG.

\subsection{Evaluation Setup}
\label{sec:eval_setup}

\noindent\textbf{Model Configuration.}
Our default model is trained with view-selective cross-attention (Stage 2) and asset-selective cross-attention (Stage 3), with Layout-Guided Structure Inversion at $t_0=0.7$ during inference.

\noindent\textbf{Compared Methods.}
While a few prior works have explored panoramic 3D scene understanding~\cite{dong2024panocontext}, their code is unavailable, making direct comparison infeasible. We therefore compare against recent single-image scene generation methods: SceneGen~\cite{meng2026scenegen}, MIDI~\cite{huang2025midi}, and SAM3D~\cite{chen2026sam}. For each test scene, we extract the perspective view that maximizes asset visibility from the ERP image by projecting 3D bounding boxes onto candidate views. This image, along with its segmentation mask~\cite{kirillov2023segment}, is provided as input. As these methods operate in their own coordinate systems, we manually align the generated scenes to the ground truth before refining with ICP~\cite{besl1992method}. Comparisons with existing methods use 20 samples due to manual alignment. Quantitative and qualitative comparisons with these methods are provided in Sec.\ref{sec:supple_4_comparison} of the supplementary.

\begin{figure}[t!]
\centering
    \includegraphics[width=1.0\textwidth]{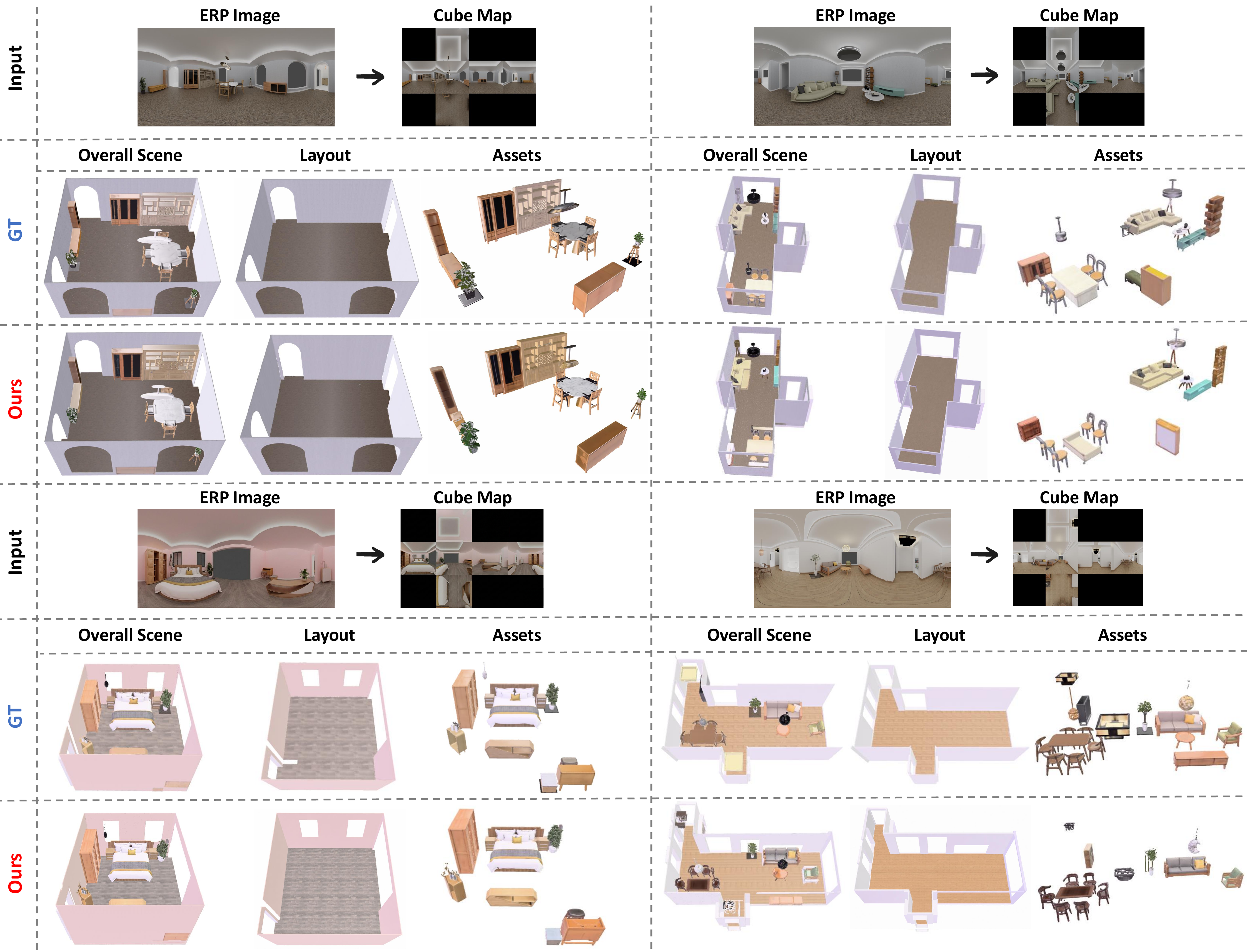}
   \caption{\textbf{Qualitative results on ERP-FRONT.} Given an ERP image and its cubemap decomposition (top), InSpace generates the overall scene, structural layout, and individual assets. Four examples are shown with ground truth for comparison.}
\label{fig:8}
\end{figure}

\begin{figure}[b!]
\centering
    \includegraphics[width=1.0\textwidth]{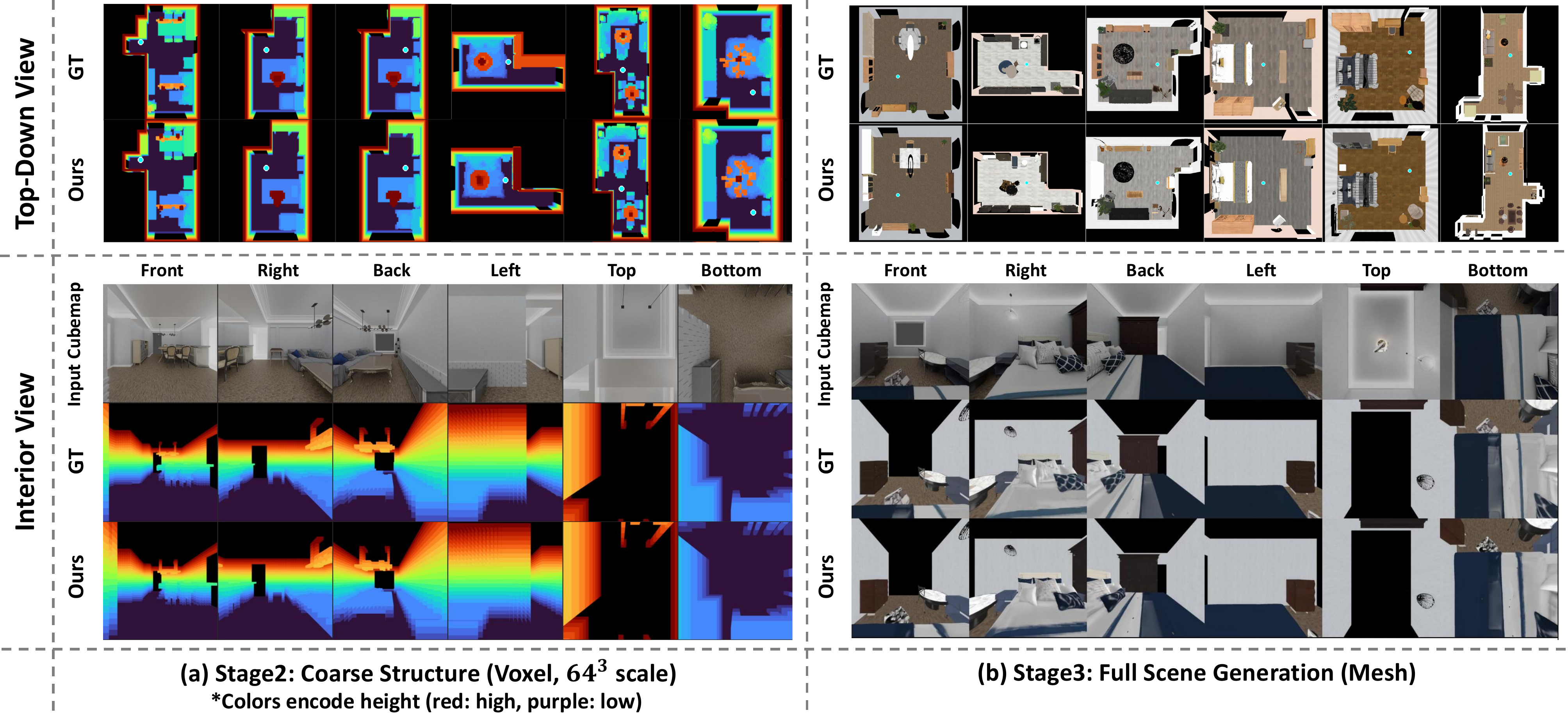}
   \caption{\textbf{Rendered evaluation views on ERP-FRONT.} (a) Coarse structure (Stage 2) and (b) full scene generation (Stage 3), rendered from a top-down view and six interior views from the camera center along cubemap directions for quantitative evaluation. Cyan dots indicate the calibrated camera center.}
\label{fig:9}
\end{figure}

\noindent\textbf{Metrics.}
For 3D evaluation, we measure Voxel IoU, Chamfer Distance (CD), and F1-Score on the coarse structure (Stage 2, Fig.~\ref{fig:9}(a)) at the scene level, as this stage captures overall room geometry before asset decomposition. On the generated meshes (Stage 3), we evaluate CD and F1-Score at both scene and asset level, where assets are matched to ground truth via estimated and GT bounding box alignment. For 2D evaluation, we render Stage~3 outputs from a top-down orthographic view and six interior views along cubemap directions (Fig.~\ref{fig:9}(b)), using PSNR and LPIPS. Both 3D and 2D metrics are also used for comparison with existing methods. 

\subsection{Results}
\label{sec:results}
For qualitative results, Fig.~\ref{fig:0_teaser} and \ref{fig:8} show generated scenes on ERP-FRONT, where InSpace produces complete scenes whose layout and assets match the ground truth. To assess generalization beyond the synthetic ERP-FRONT, we further test on ReplicaPano~\cite{dong2024panocontext}, which provides ERP images extracted from the realistic Replica dataset~\cite{straub2019replica}. Additional results on ERP-FRONT and ReplicaPano, along with comparisons against single-image scene generation methods, are provided in Sec.\ref{sec:supple_3_qualitative} of the supplementary.
For quantitative results, Table~\ref{tab:results} shows performance under different configurations. $t_0\!=\!0.5$ achieves the best IoU in Stage~2, while Stage~3 performance remains largely consistent across $t_0$ values, with $t_0\!=\!0.7$ performing marginally better across most metrics. This difference in the optimal $t_0$ arises because the two stages evaluate different levels of geometric detail: Stage~2 measures coarse structure on a sparse $64^3$ grid, whereas Stage~3 measures the final mesh via dense point sampling, and favor slightly different $t_0$.

\begin{table}[t!]
\centering
\caption{\textbf{Quantitative results on ERP-FRONT test set.} (Left) Coarse structure generation (Stage~2) under different configurations. VS: view-selective cross-attention. Inv: Layout-Guided Structure Inversion. (Right) Full scene generation (Stage~3) at scene and asset level. All CD values are multiplied by $10^3$. \colorbox{best}{\smash{Best}} and \colorbox{second}{\smash{second-best}} are highlighted.}
\label{tab:results}
\resizebox{\linewidth}{!}{%
\begin{tabular}{lccccc|lccccccc}
\toprule
& \multicolumn{5}{c|}{\textbf{Stage 2: Coarse Structure}} & & \multicolumn{7}{c}{\textbf{Stage 3: Full Scene Generation}} \\
\cmidrule(lr){2-6} \cmidrule(lr){8-14}
& & & & & & & & & & \multicolumn{2}{c}{PSNR$\uparrow$} & \multicolumn{2}{c}{LPIPS$\downarrow$} \\
\cmidrule(lr){11-12} \cmidrule(lr){13-14}
Configuration & IoU$\uparrow$ & F1$\uparrow$ & Prec.$\uparrow$ & Rec.$\uparrow$ & \makecell{CD$\downarrow$\\{\scriptsize($\times 10^3$)}} & Level & \makecell{CD$\downarrow$\\{\scriptsize($\times 10^3$)}} & F1@.01$\uparrow$ & F1@.02$\uparrow$ & \makecell{Top\\{-down}} & \makecell{Int\\{-erior}} & \makecell{Top\\{-down}} & \makecell{Int\\{-erior}} \\
\midrule
\rowcolor{subheader}
\multicolumn{6}{l}{\textit{Trained w/o VS-CrossAttn}}  \\
w/o VS & 44.36 & 55.61 & 55.69 & 55.76 & 15.92 & & - & - & - & - & - & - & - \\
\midrule
\rowcolor{subheader}
\multicolumn{6}{l}{\textit{Trained w/ VS-CrossAttn}} & \multicolumn{8}{l}{\cellcolor{subheader_as}\textit{Trained w/ AS-CrossAttn}} \\
w/ VS ($t_0\!=\!1.0$) & 57.54 & 69.07 & 69.19 & 69.12 & 7.68 & Scene & 1.52 & 35.26 & 79.89 & 19.02 & 11.22 & 0.228 & 0.653 \\
 & & & & & & Asset & 1.92 & 53.42 & 76.28 & - & - & - & - \\
\quad + Inv ($t_0\!=\!0.3$) & \cellcolor{second}58.06 & \cellcolor{second}69.66 & \cellcolor{second}70.07 & \cellcolor{second}69.49 & \cellcolor{best}7.41 & Scene & 1.48 & \cellcolor{second}36.63 & 81.48 & \cellcolor{best}19.17 & 11.22 & \cellcolor{best}0.219 & \cellcolor{second}0.651 \\
 & & & & & & Asset & 1.80 & \cellcolor{second}55.42 & \cellcolor{second}78.65 & - & - & - & - \\
\quad + Inv ($t_0\!=\!0.5$) & \cellcolor{best}58.48 & \cellcolor{best}69.97 & \cellcolor{best}70.33 & \cellcolor{best}69.82 & \cellcolor{second}7.66 & Scene & \cellcolor{second}0.79 & 36.01 & \cellcolor{second}81.69 & \cellcolor{second}19.16 & 11.22 & 0.224 & 0.654 \\
 & & & & & & Asset & \cellcolor{best}1.02 & 53.77 & 77.59 & - & - & - & - \\
\quad + Inv ($t_0\!=\!0.7$) & 57.09 & 68.78 & 68.97 & 68.79 & 7.76 & Scene & \cellcolor{best}0.63 & \cellcolor{best}36.69 & \cellcolor{best}82.10 & \cellcolor{best}19.17 & 11.22 & \cellcolor{second}0.220 & \cellcolor{best}0.650 \\
 & & & & & & Asset & \cellcolor{second}1.72 & \cellcolor{best}56.71 & \cellcolor{best}79.52 & - & - & - & - \\

\bottomrule
\end{tabular}%
}
\end{table}

\subsection{Ablation Study}
\label{sec:ablation}
Stage~2 generates the coarse scene structure that determines the overall geometry, while Stage~3 refines its details. Thus, the fundamental layout is largely established in Stage~2, making its design choices critical. To evaluate the effectiveness of view-selective (VS) cross-attention, we train an additional model with standard cross-attention (w/o VS) alongside our default model (w/ VS).

\begin{figure}[t!]
\centering
    \includegraphics[width=\textwidth]{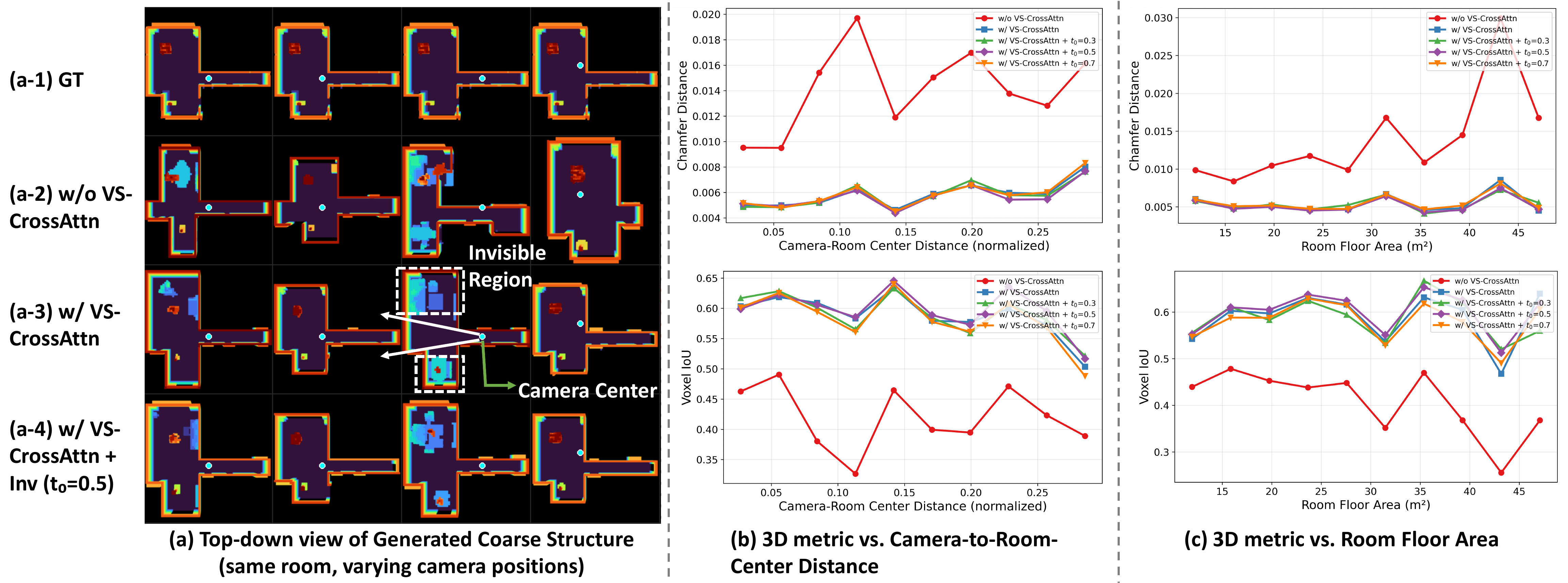}
    \caption{\textbf{Ablation on Stage~2 across camera positions.} (a) Generated coarse structure from different camera positions in the same room. (b, c) Quantitative metrics across multiple rooms: w/o VS degrades sharply with camera-to-room-center distance (b) and room floor area (c), while VS with Inversion remains robust.}
\label{fig:10}
\end{figure}

\noindent\textbf{View-Selective Cross-Attention.}
Table~\ref{tab:results} (left) compares two models trained with and without view-selective (VS) cross-attention on Stage 2. Without VS, every voxel is conditioned on the full cubemap, regardless of which faces are visible from its position, yielding an IoU of 44.36. With inversion disabled for both ($t_0\,{=}\,1.0$), VS raises IoU to 57.54 and nearly halves CD ($15.92 \rightarrow 7.68$). This confirms the importance of spatial conditioning. Without VS, each voxel is influenced by conflicting cues from all six faces depicting disjoint parts of the room, whereas restricting attention to the visible faces removes this ambiguity and yields geometry consistent with the room shape (Fig.~\ref{fig:10}(a-2 vs.\ a-3)).

\noindent\textbf{Layout-Guided Structure Inversion.}
Using the w/ VS model, we vary $t_0$ at inference (Table~\ref{tab:results}). For Stage 2, $t_0\!=\!0.5$ achieves the best IoU, while for Stage 3, $t_0\!=\!0.7$ is marginally better across most metrics. Since $t_0$ controls only the initialization of Stage 2, it produces a single coarse structure (CSG) that both stages inherit. The two stages, however, evaluate it under different criteria. Stage 2 measures coarse occupancy on a sparse $64^3$ voxel grid, where a stronger prior (lower $t_0$) best matches the ground-truth voxels. Stage 3 instead measures the final mesh via dense point sampling, where a slightly weaker prior (higher $t_0$) is preferable. Because the PSG covers only the visible surfaces, relaxing the prior lets the model complete occluded regions more evenly, yielding a CSG that fills in details absent from the sparse Stage 2 occupancy and that the mesh stage can refine into finer geometry. Overall, all inversion settings yield consistently strong results with marginal differences, confirming the robustness of Layout-Guided Structure Inversion.

\noindent\textbf{Camera Position Robustness.}
Fig.~\ref{fig:10}(a) shows coarse structures from different camera positions in the same room on Stage 2. The w/o VS model predicts inconsistent and incorrect layouts across positions (a-2), while the w/ VS model consistently recovers accurate room layouts (a-3,a-4). Notably, when the camera position introduces occluded regions, the model often generates assets in invisible areas. Fig.~\ref{fig:10}(b) measures Voxel IoU and CD as a function of camera-to-room-center distance. The w/ VS model maintains performance, while the w/o VS model degrades rapidly. 
Fig.~\ref{fig:10}(c) shows that the w/ VS model maintains robust performance across room sizes, while the w/o VS model degrades rapidly. This gap widens with room size because, as the room grows, the cubemap faces cover increasingly disjoint regions, so attending to all of them forces each voxel to reconcile cues from distant, unrelated areas. VS instead confines conditioning to the locally visible faces and thus remains robust.

\section{Conclusion}
We presented InSpace, a framework for generating complete 3D indoor scenes from a single ERP image. Leveraging full 360° coverage, InSpace introduces view-selective and asset-selective cross-attention for spatially grounded generation, with Layout-Guided Structure Inversion to incorporate geometric priors at inference.
To enable training and evaluation, we constructed ERP-FRONT, a paired ERP-Image-to-3D indoor scene dataset based on 3D-FRONT. Experiments show that InSpace achieves strong performance across both 3D and 2D metrics on ERP-FRONT, and generalizes beyond synthetic data to realistic panoramic scenes such as ReplicaPano.

\section*{Acknowledgements}
This work was supported by the Korea Planning \& Evaluation Institute of Industrial Technology(KEIT) and the Ministry of Trade, Industry \& Resources(MOTIR) of the Republic of Korea (RS-2024-00417108), and supported by the Institute for Information \& communications Technology Planning \& Evaluation (IITP) grant funded by the Korea government(MSIT) (No.RS-2021-II211381, Development of Causal AI through Video Understanding and Reinforcement Learning, and Its Applications to Real Environments).

\bibliographystyle{splncs04}
\bibliography{main}

\input{supplementary}

\end{document}

%% file: supplementary.tex
% ============================================================
% Supplementary Material — BODY ONLY (\input from main.tex)
% arXiv merged-PDF version. Sections A, B, C ... (\appendix);
% figures/tables/equations numbered within each section (A.1, B.2, ...).
% Required in main.tex preamble: tabularx, array, enumitem,
%   \definecolor{accentblue}{RGB}{25,55,109}, amsmath (for \numberwithin).
% ============================================================
\clearpage
\appendix
% \appendix auto-resets the section counter and renders sections as A, B, C, ...
% (do NOT add \setcounter{section}{0} or \renewcommand{\thesection} here)
\setcounter{figure}{0}
\setcounter{table}{0}
\setcounter{equation}{0}
\numberwithin{figure}{section}
\numberwithin{table}{section}
\numberwithin{equation}{section}
% ---------------------------------------------------------------
% IMPORTANT (hyperref): give the supplementary UNIQUE anchor names.
% Resetting counters above makes supp anchors collide with the main
% paper's (both become "section.1", "figure.1", ...), so clicking a
% supp link (ToC or \ref) would jump into the MAIN paper. Prefixing
% every supp anchor with "supp." prevents that.
% ---------------------------------------------------------------
\renewcommand{\theHsection}{supp.\thesection}
\renewcommand{\theHsubsection}{supp.\thesubsection}
\renewcommand{\theHsubsubsection}{supp.\thesubsubsection}
\renewcommand{\theHfigure}{supp.\thefigure}
\renewcommand{\theHtable}{supp.\thetable}
\renewcommand{\theHequation}{supp.\theequation}

% --- Lightweight supplementary title (author list intentionally NOT repeated) ---
\begin{center}
  {\Large\bfseries Supplementary Material for\\[2pt]
   InSpace: Structure-Aware 3D Indoor Scene Generation from a Single 360° Image\par}
\end{center}
% \vspace{8pt}

\newcommand{\tocentry}[4]{%
  \noindent\begin{minipage}{\linewidth}
    {\color{accentblue}\textbf{#1}}\hspace{0.6em}%
    {\textbf{\hyperref[#2]{\textcolor{black}{#3}}}}\hfill{\small\hyperref[#2]{\textcolor{red}{p.\pageref*{#2}}}}\par
    \vspace{2pt}
    \hspace{2em}\begin{minipage}{\dimexpr\linewidth-2em}
      {\small\color{gray!70!black} #4}
    \end{minipage}
  \end{minipage}\par
  \vspace{7pt}
}

\begin{figure*}[h!]
% \captionsetup{skip=5pt}
\renewcommand{\thefigure}{A.\arabic{figure}}   % section=0이라 letter가 비니까 A 강제
\begin{center}
\vskip -0.3in
\centerline{\includegraphics[width=\textwidth]{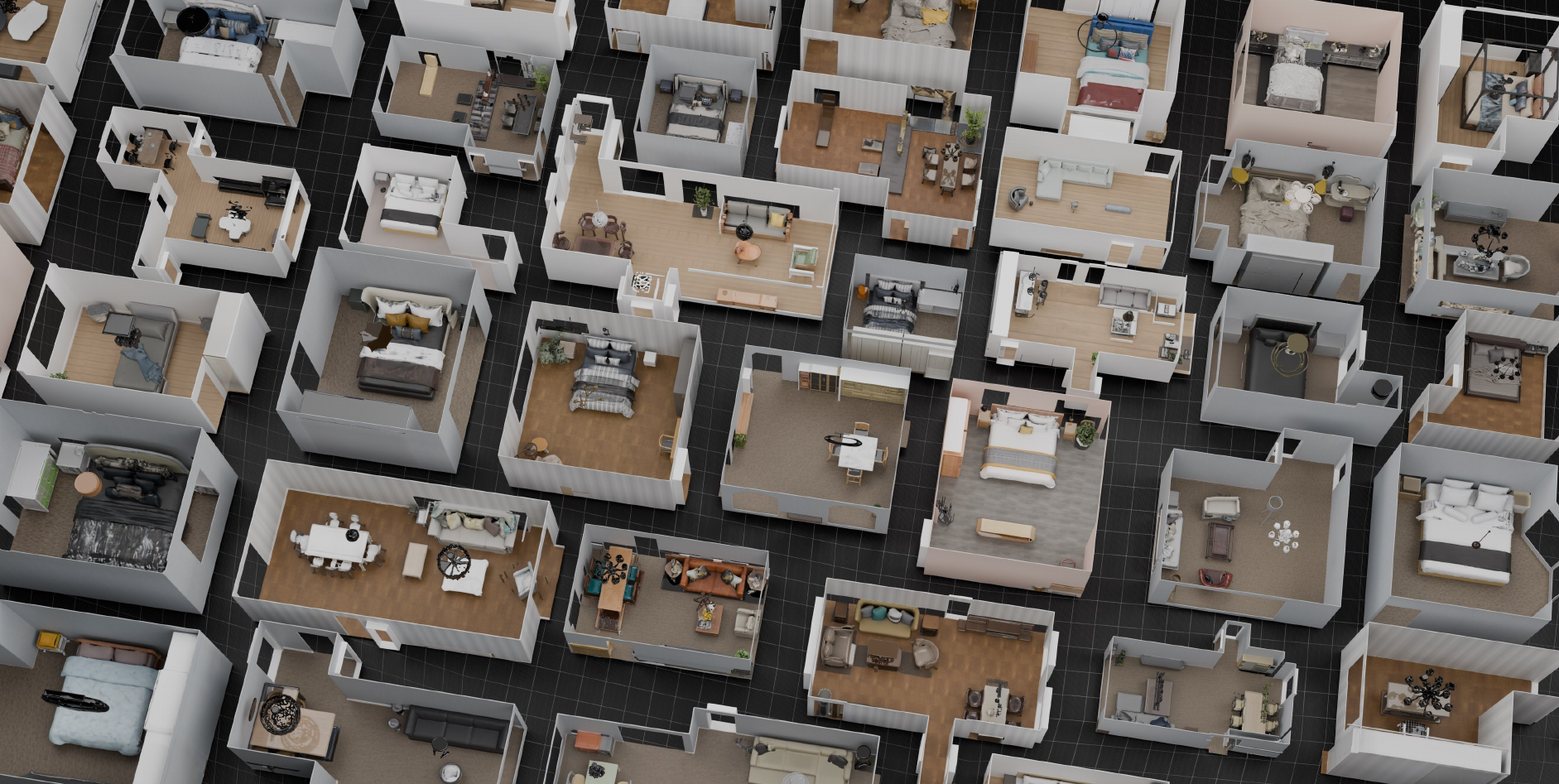}}
\vskip -0.1in
\caption{InSpace generates complete 3D indoor scenes from a single 360° image.}
\label{fig:0_teaser}
\end{center}
\vskip -0.4in
\end{figure*}

\noindent This supplementary material provides additional implementation details, dataset statistics, and experimental results that complement the main paper. For ease of navigation, the contents are organized as summarized below, with direct links to the corresponding sections.

\vspace{8pt}
\noindent{\color{accentblue}\rule{\linewidth}{1.5pt}}\par
\vspace{1pt}
\begin{center}
  {\small\color{accentblue}\textbf{Supplementary Contents}}
\end{center}
\vspace{-5pt}
\noindent{\color{accentblue}\rule{\linewidth}{0.4pt}}\par
\vspace{8pt}
\tocentry{A}{sec:supple_1_erpfront}{ERP-FRONT Dataset}%
  {Dataset construction pipeline, camera placement strategy, and detailed dataset statistics.}
\tocentry{B}{sec:supple_2_bbox}{3D Bounding Box Estimator}%
  {Model architecture, qualitative and quantitative results of the 3D OBB detector.}
\tocentry{C}{sec:supple_3_qualitative}{More Qualitative Results}%
  {Additional generated scenes on ERP-FRONT and ReplicaPano datasets.}
\tocentry{D}{sec:supple_4_comparison}{Comparison with Single-Image to Scene Generation Methods}%
  {Qualitative and quantitative comparisons with SceneGen, MIDI, and SAM3D.}
\tocentry{E}{sec:supple_6_limitation}{Limitation and Future Work}%
  {Discussion of current limitations and future directions.}
\vspace{3pt}
\noindent{\color{accentblue}\rule{\linewidth}{0.4pt}}\par
\vspace{1pt}
\noindent{\color{accentblue}\rule{\linewidth}{1.5pt}}\par
\vspace{10pt}

\clearpage

\section{ERP-FRONT Dataset}
\label{sec:supple_1_erpfront}
\setcounter{figure}{1} 

\subsection{Overview}
ERP-FRONT is a synthetic dataset of paired equirectangular projection (ERP) images and 3D indoor scene meshes, constructed from 3D-FRONT \cite{fu20213d-3d-front} and 3D-FUTURE \cite{fu20213d-3d-future} datasets, which provide high-quality furnished indoor scenes with semantic annotations.

\subsubsection{3D-FRONT and 3D-FUTURE.}
The original 3D-FRONT \cite{fu20213d-3d-front} dataset is organized at the house level, where each house consists of one or more rooms connected into a single scene. 3D-FRONT provides the structural definition of each room, including layout elements such as walls, floors, and ceilings, along with scene configuration specifying the position and orientation of each furniture piece within the room. The actual 3D asset meshes (furniture and objects) come from the 3D-FUTURE  \cite{fu20213d-3d-front} dataset, which contains a large collection of high-quality textured furniture models. In short, 3D-FRONT defines the room structure and asset arrangement, while 3D-FUTURE supplies the corresponding asset geometry and appearance.

\begin{figure}[t!]
\centering
    \includegraphics[width=1.0\textwidth]{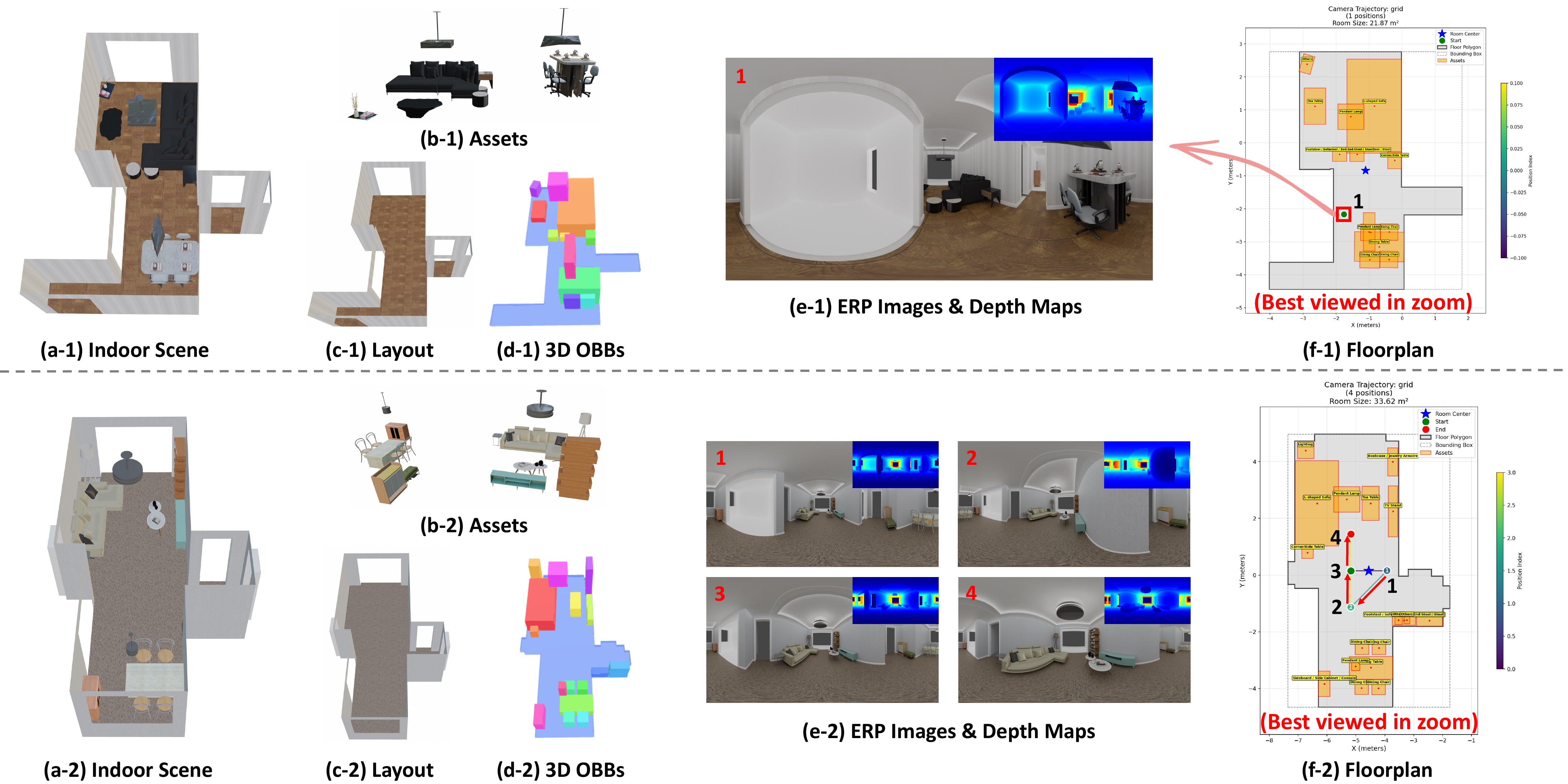}
   \caption{\textbf{Example scenes from ERP-FRONT.} Two rooms are shown (top and bottom rows). Each scene includes: (a) complete scene mesh, (b) individual asset meshes, (c) structural layout mesh, (d) 3D oriented bounding boxes (OBBs) for assets, (e) ERP images with depth maps rendered from valid camera positions, and (f) a 2D floorplan with the floor polygon, asset bounding boxes, and camera trajectory.}
\label{fig:supple_1}
\end{figure}

\subsubsection{Scene Components.}
Since InSpace targets indoor scene generation at the room level, we construct ERP-FRONT by processing each room as an independent scene. Each scene comprises six components, as illustrated in Fig. \ref{fig:supple_1}.
\begin{enumerate}
    \item \textbf{Complete indoor scene mesh} (Fig. \ref{fig:supple_1}(a)): the full room geometry combining both the structural layout and all placed assets.
    \item \textbf{Individual asset meshes} (Fig. \ref{fig:supple_1}(b)): each furniture or object mesh from 3D-FUTURE placed within the room at the position and orientation specified by 3D-FRONT.
    \item \textbf{Structural layout mesh} (Fig. \ref{fig:supple_1}(c)): the structural surfaces of the room (floors and walls), extracted from the 3D-FRONT structural annotations.
    \item \textbf{3D oriented bounding boxes} (Fig. \ref{fig:supple_1}(d)): a 3D OBB for each asset, computed from the asset's mesh vertices.
    \item \textbf{ERP images with depth maps} (Fig. \ref{fig:supple_1}(e)): one or more equirectangular panoramic images rendered from valid camera positions within the room, each paired with a corresponding equirectangular depth map.
    \item \textbf{Floorplan} (Fig. \ref{fig:supple_1}(f)): a 2D top-down map of the room including the floor polygon, asset bounding boxes, and camera positions.
\end{enumerate}
The complete scene mesh, structural layout mesh, and all individual asset meshes are all defined in the same room coordinate system as specified in 3D-FRONT, so that combining them directly reconstructs the complete indoor scene.

\subsection{Dataset Statistics.}
\begin{figure}[t!]
\centering
    \includegraphics[width=1.0\textwidth]{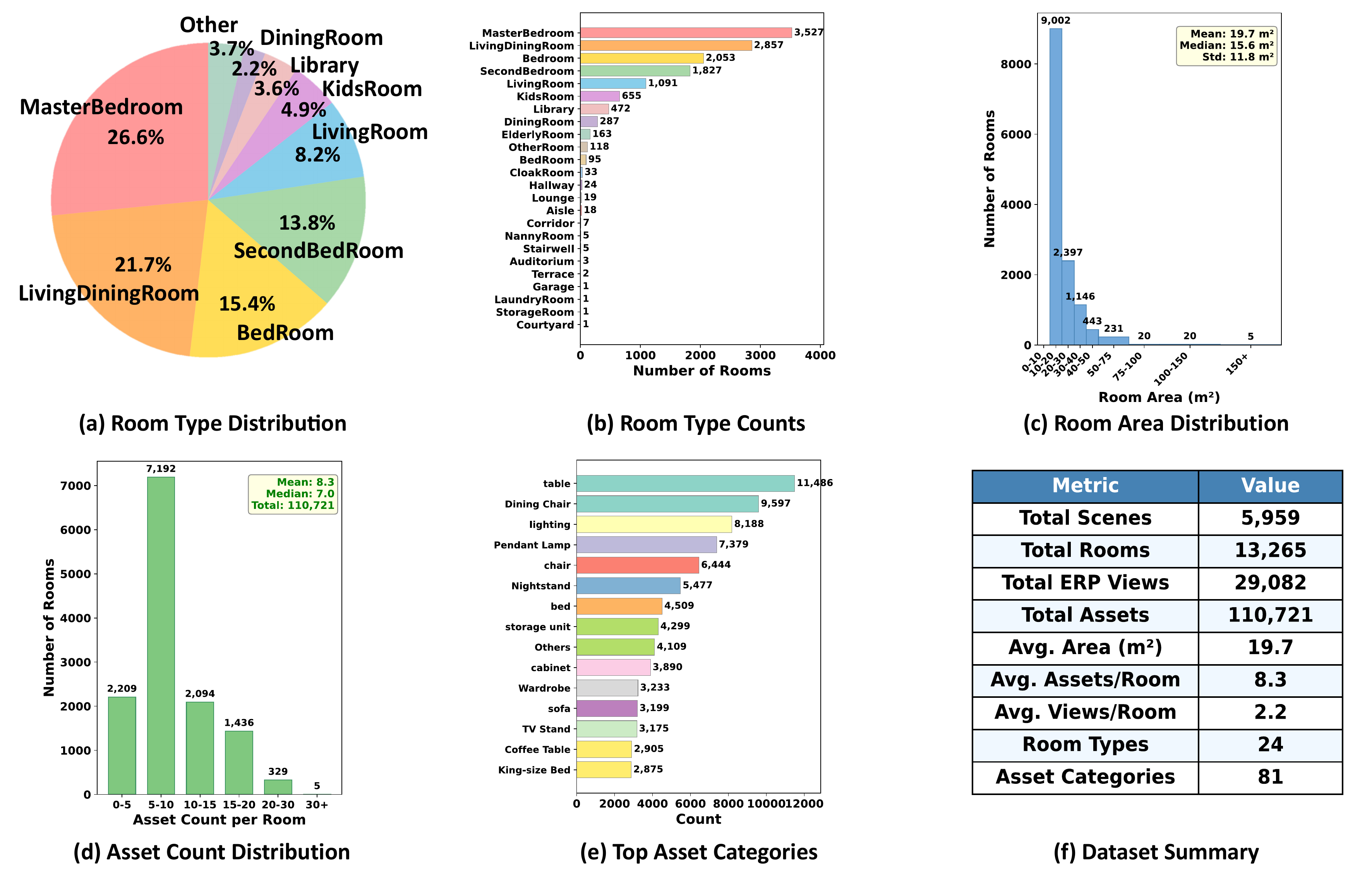}
   \caption{\textbf{ERP-FRONT dataset statistics.} 
    (a) Distribution of room types in the dataset. 
    (b) Number of rooms per type. 
    (c) Distribution of room areas. 
    (d) Distribution of asset counts per room. 
    (e) Most frequent asset categories by instance count. 
    (f) Overall dataset summary.}
\label{fig:supple_2}
\end{figure}

ERP-FRONT comprises 5,959 scenes across 13,265 rooms, yielding about 30k (29,082) ERP image–mesh pairs (Fig.~\ref{fig:supple_2}(f)). The dataset includes 24 room types, with bedrooms and living–dining spaces being the most common (Fig.~\ref{fig:supple_2} (a,b)). Room sizes vary substantially, with an average floor area of 19.7\,m$^2$, reflecting diverse indoor layouts (Fig.~\ref{fig:supple_2}(c)). On average, each room contains 8.3 assets and 2.2 rendered ERP views, where the number of valid camera positions depends on room size and furniture density (Fig.~\ref{fig:supple_2}(d)). The dataset also includes a wide range of furniture categories such as tables, chairs, and lighting, capturing the diversity of furnished indoor environments in 3D-FRONT (Fig.~\ref{fig:supple_2}(e)).

%-------------------------------------------------------------------------------
\subsection{Dataset Construction}
\label{subsec:dataset_construction}
%-------------------------------------------------------------------------------

\subsubsection{ERP Image Rendering.}
We render all ERP images using BlenderProc~\cite{denninger2019blenderproc}, a scriptable physically-based rendering pipeline built on Blender. The pipeline proceeds in three stages: (i)~\emph{Room assembly}, where the structural layout and all 3D-FUTURE assets are loaded according to the 3D-FRONT scene configuration; (ii)~\emph{Camera placement}, where valid panoramic camera locations are determined for each room (described below); and (iii)~\emph{Rendering}, where ERP color images and depth maps are generated from each valid camera position.

\subsubsection{Camera Placement via Grid Sampling.}
To determine valid ERP camera locations, we uniformly sample candidate positions on a 2D grid over the floor plane (Fig. \ref{fig:supple_1} (f)). Sampling is restricted to the central 70\% of the floor area with a grid spacing of $1.5\,\text{m}$ and an overlap margin of $0.3\,\text{m}$. 
This restriction avoids placing cameras too close to room boundaries, where panoramic views often contain limited spatial context due to severe wall occlusions. It also better reflects realistic capture setups, where panoramic cameras are typically positioned away from corners and large obstacles. Furthermore, placing a camera directly on or inside scene objects would lead to severe occlusions and unstable observations in the rendered ERP images. To prevent this, we discard candidates that fall within a $0.3\,\text{m}$ margin of any asset's 2D bounding box.
The camera height is fixed at $1.159\,\text{m}$ above the floor for all positions. Only candidates satisfying the clearance constraint are retained as valid camera locations. If no valid candidate exists for a given room, we fall back to selecting a single position near the room center even if it slightly overlaps with assets, ensuring that every room yields at least one ERP observation. This procedure produces on average $\sim$2.2 camera positions per room, providing diverse viewpoints and camera-to-room-center distances across ERP images of the same scene.

\section{3D Bounding Box Estimator}
\label{sec:supple_2_bbox}

\subsection{Model Architecture}
The 3D bounding box estimator takes the Coarse Scene Geometry (CSG), a binary occupancy grid of resolution $64^3$, as input and predicts 7-DoF oriented bounding boxes $\{(\mathbf{o}^c_k, \mathbf{o}^s_k, \theta_k)\}$ for each asset in a single forward pass, where $\mathbf{o}^c_k \in [-0.5, 0.5]^3$ is the center, $\mathbf{o}^s_k \in [0, 0.5]^3$ is the extent, and $\theta_k \in [-\pi, \pi]$ is the yaw angle.
Since the CSG is a low-resolution ($64^3$) voxel grid, it is essential to preserve fine-grained 3D spatial structure throughout detection. We therefore adopt a 3D U-Net backbone, which preserves full spatial resolution via encoder–decoder skip connections. This allows the detector to leverage fine-grained 3D structural cues for accurate object localization. Inspired by CenterPoint~\cite{yin2021center}, we further adopt a dense, anchor-free detection paradigm, where objects are represented by their center points on the voxel grid rather than predefined anchor boxes.
The backbone first projects the CSG $\in \{0,1\}^{1 \times 64^3}$ to a 32-channel feature volume via an initial Conv3d to encode sufficient spatial context from the sparse binary input, then processes it through four encoders with a symmetric decoder and skip connections, producing a full-resolution feature map $\mathbf{F} \in \mathbb{R}^{32 \times 64^3}$. 
Object detection is performed by identifying peaks in a \emph{heatmap} \cite{yin2021center}: a dense probability map over all $64^3$ voxels. During training, each GT center is rendered as a 3D Gaussian on the heatmap, so that voxels near the center receive graded soft supervision proportional to their distance, rather than a hard binary label.
Five lightweight heads then independently predict object attributes from $\mathbf{F}$:

\begin{enumerate}[leftmargin=1.5em, itemsep=1pt, topsep=2pt]
    \item \textbf{Heatmap head}: predicts a per-voxel center confidence map $\in [0,1]^{1 \times 64^3}$ via sigmoid activation, where each scalar indicates the probability of an object center at that location.
    \item \textbf{Offset head}: regresses sub-voxel offsets to the precise object center $\mathbf{o}^c_k \in \mathbb{R}^3$.
    \item \textbf{Size head}: regresses bounding box extents $\mathbf{o}^s_k \in \mathbb{R}^3$.
    \item \textbf{Rotation bin head}: classifies yaw angle $\theta_k \in \mathbb{R}$ into one of 12 discrete bins covering $[-\pi, \pi]$.
    \item \textbf{Rotation residual head}: regresses fine-grained residuals $\in \mathbb{R}^{12}$ within the predicted bin.
\end{enumerate}
The final detections are obtained by extracting heatmap peaks followed by IoU-based NMS.

\begin{figure}[t!]
\centering
    \includegraphics[width=1.0\textwidth]{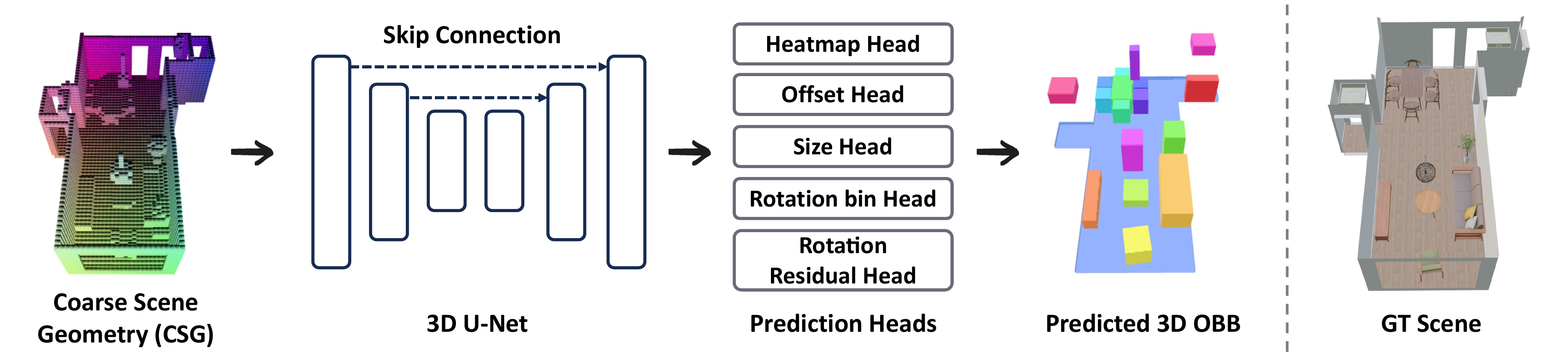}
   \caption{\textbf{3D Bounding Box Estimator.} The Coarse Scene Geometry (CSG) is first projected to a 32-channel feature volume via Conv3d, then fed into a 3D U-Net with skip connections to produce a full-resolution feature map. Five prediction heads then take this feature map as input and estimate 3D oriented bounding boxes (OBBs) for each asset. Predicted OBBs (colored boxes) are compared against the GT scene (right).}
\label{fig:supple_3}
\end{figure}

\begin{figure}[t!]
\centering
    \includegraphics[width=1.0\textwidth]{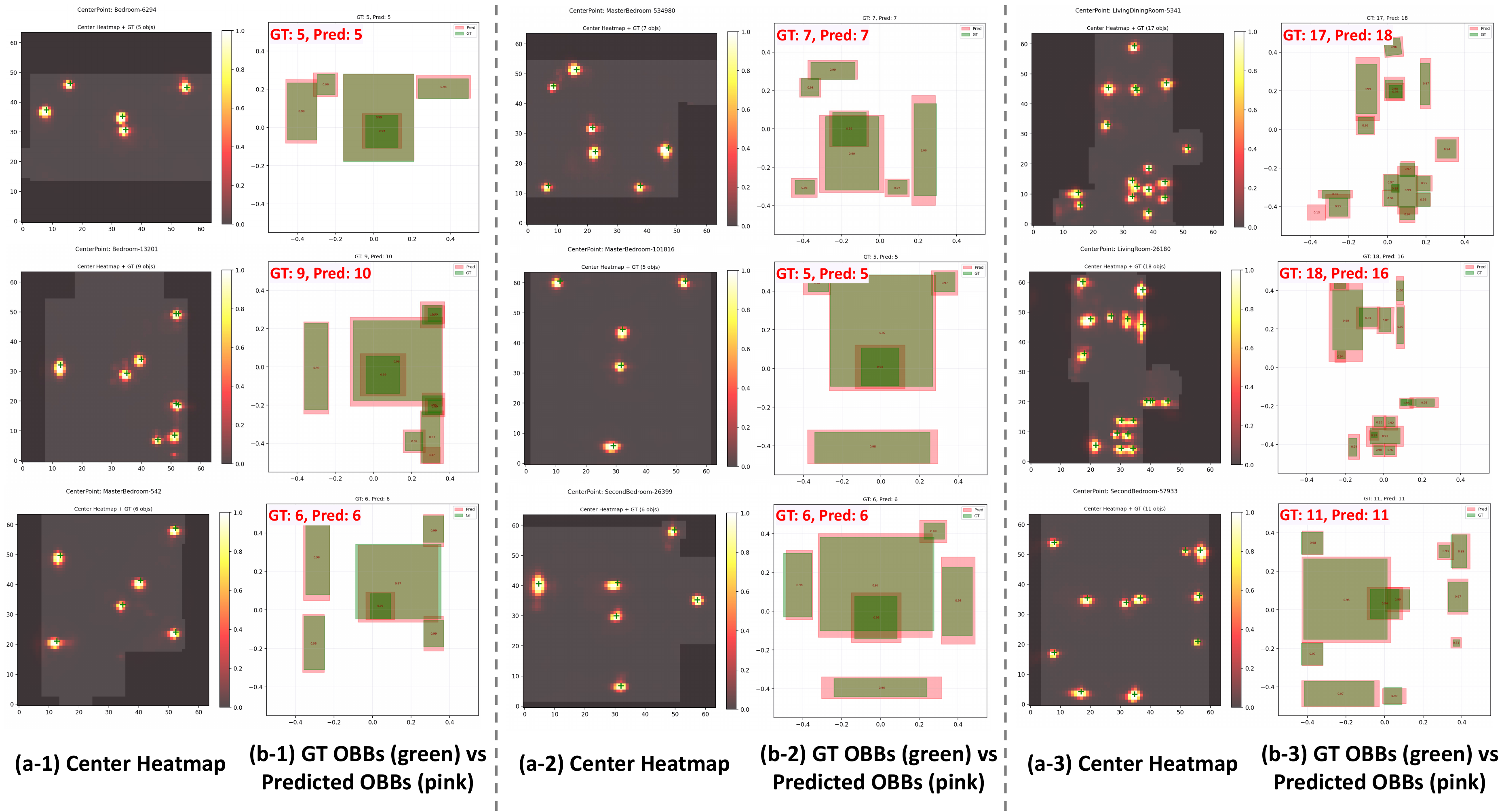}
   \caption{\textbf{Qualitative Results of the 3D Bounding Box Estimator on ERP-FRONT.} Each example shows (a) the predicted center heatmap (top-down view of the $64^3$ voxel grid, brighter regions indicate higher center confidence) and (b) predicted OBBs (pink) overlaid with GT OBBs (green), with GT and predicted object counts annotated in red (top-left of each panel). The heatmap peaks closely align with GT object centers across diverse room types and object counts.}
\label{fig:supple_4}
\end{figure}

\subsection{Training Loss}
The model is trained end-to-end with a composite loss over all five heads:
\begin{equation}
    \mathcal{L} = \lambda_{\text{hm}}\mathcal{L}_{\text{hm}}
                + \lambda_{\text{off}}\mathcal{L}_{\text{off}}
                + \lambda_{\text{sz}}\mathcal{L}_{\text{sz}}
                + \lambda_{\text{rb}}\mathcal{L}_{\text{rb}}
                + \lambda_{\text{rr}}\mathcal{L}_{\text{rr}}
                + \lambda_{\text{dr}}\mathcal{L}_{\text{dr}}.
\end{equation}
The heatmap loss $\mathcal{L}_{\text{hm}}$ applies a Gaussian focal loss~\cite{law2018cornernet} over all $64^3$ voxels, which down-weights easy negatives and focuses training on object center regions. The offset loss $\mathcal{L}_{\text{off}}$ and size loss $\mathcal{L}_{\text{sz}}$ are L1 losses computed only at GT center voxels. The rotation bin loss $\mathcal{L}_{\text{rb}}$ is a cross-entropy loss for coarse bin classification, and the rotation residual loss $\mathcal{L}_{\text{rr}}$ is a Smooth L1 loss for within-bin refinement. Finally, $\mathcal{L}_{\text{dr}}$ provides auxiliary dense rotation supervision over voxels within the Gaussian radius of each GT center, weighted by their heatmap value. Loss weights are detailed in Sec.~\ref{sec:supple_2_impl}.
\subsection{Implementation Details}
\label{sec:supple_2_impl}
The detector consists of a 3D U-Net with four encoder stages of channels 32 → 64 → 128 → 256 (stride-2 convolutions) and a symmetric decoder, totaling 12.45M parameters. It is trained on the ERP-FRONT training split (12,120 rooms) for 45 epochs with batch size 32, using AdamW (lr=$10^{-4}$, weight decay=$10^{-4}$) with cosine annealing. Loss weights are set to $(\lambda_{\text{hm}}, \lambda_{\text{off}}, \lambda_{\text{sz}}, \lambda_{\text{rb}}, \lambda_{\text{rr}}, \lambda_{\text{dr}}) = (1, 1, 5, 2, 2, 1)$, with $\lambda_{\text{sz}}$ weighted highest as accurate extents are critical for downstream asset generation. At inference, object centers are extracted as local maxima of the heatmap via 3D max-pool NMS (kernel $7^3$, threshold 0.3), followed by IoU-based NMS (threshold 0.3) to remove duplicates, keeping up to 50 detections per scene.

\subsection{Results}

\noindent\textbf{Qualitative Results.}
Fig.~\ref{fig:supple_4} shows qualitative results on the ERP-FRONT test set. The predicted center heatmap (a) shows sharp, well-localized peaks that closely align with GT object centers across diverse room types and object counts. The predicted OBBs (b, pink) largely overlap with GT OBBs (green), confirming accurate localization. While box sizes and object counts occasionally deviate from GT, the predicted OBBs provide sufficiently accurate center positions and extents to enable downstream asset generation at the correct locations with plausible sizes.

\noindent\textbf{Quantitative Results.}
Table~\ref{tab:bbox} shows detection performance on the ERP-FRONT test set. We evaluate using 3D and 2D OBB metrics at three IoU thresholds (0.25, 0.50, 0.75), along with mean center error. The model achieves strong performance at IoU@0.25 (F1: 85.3\% in 3D, 87.0\% in 2D) and reasonable performance at IoU@0.50. The drop at IoU@0.75 is expected, as 3D OBB IoU is highly sensitive to small errors in size or yaw angle, where even minor misalignments cause a steep reduction in volumetric overlap. Importantly, the mean center error of 6.5\,cm demonstrates precise object localization, enabling accurate spatial grounding of assets in the downstream generation stages.

\begin{table}[t!]
\centering
\caption{Quantitative results of the 3D bounding box estimator on the ERP-FRONT test set.
Recall, Precision, and F1 are reported at IoU thresholds of 0.25, 0.50, and 0.75 
for both 3D and 2D (bird's-eye view) OBB metrics.}
\label{tab:bbox}
\small
\setlength{\tabcolsep}{6pt}
\renewcommand{\arraystretch}{1.2}
\begin{tabularx}{\linewidth}{l *{3}{>{\centering\arraybackslash}X} *{3}{>{\centering\arraybackslash}X}}
\toprule
& \multicolumn{3}{c}{\textbf{3D OBB}} & \multicolumn{3}{c}{\textbf{2D OBB (BEV)}} \\
\cmidrule(lr){2-4} \cmidrule(lr){5-7}
& @0.25 & @0.50 & @0.75 & @0.25 & @0.50 & @0.75 \\
\midrule
Recall    & 90.6 & 79.4 & 29.3 & 91.9 & 86.6 & 55.1 \\
Precision & 88.3 & 77.4 & 28.5 & 89.6 & 84.4 & 53.6 \\
F1-Score  & 89.5 & 78.4 & 28.9 & 90.7 & 85.5 & 54.3 \\
\midrule
Mean IoU        & \multicolumn{3}{c}{67.2}      & \multicolumn{3}{c}{75.3} \\
Center Error    & \multicolumn{3}{c}{6.3 cm}    & \multicolumn{3}{c}{4.4 cm} \\
\bottomrule
\end{tabularx}
\end{table}

\begin{figure}[p]
\centering
    \includegraphics[width=1.0\textwidth]{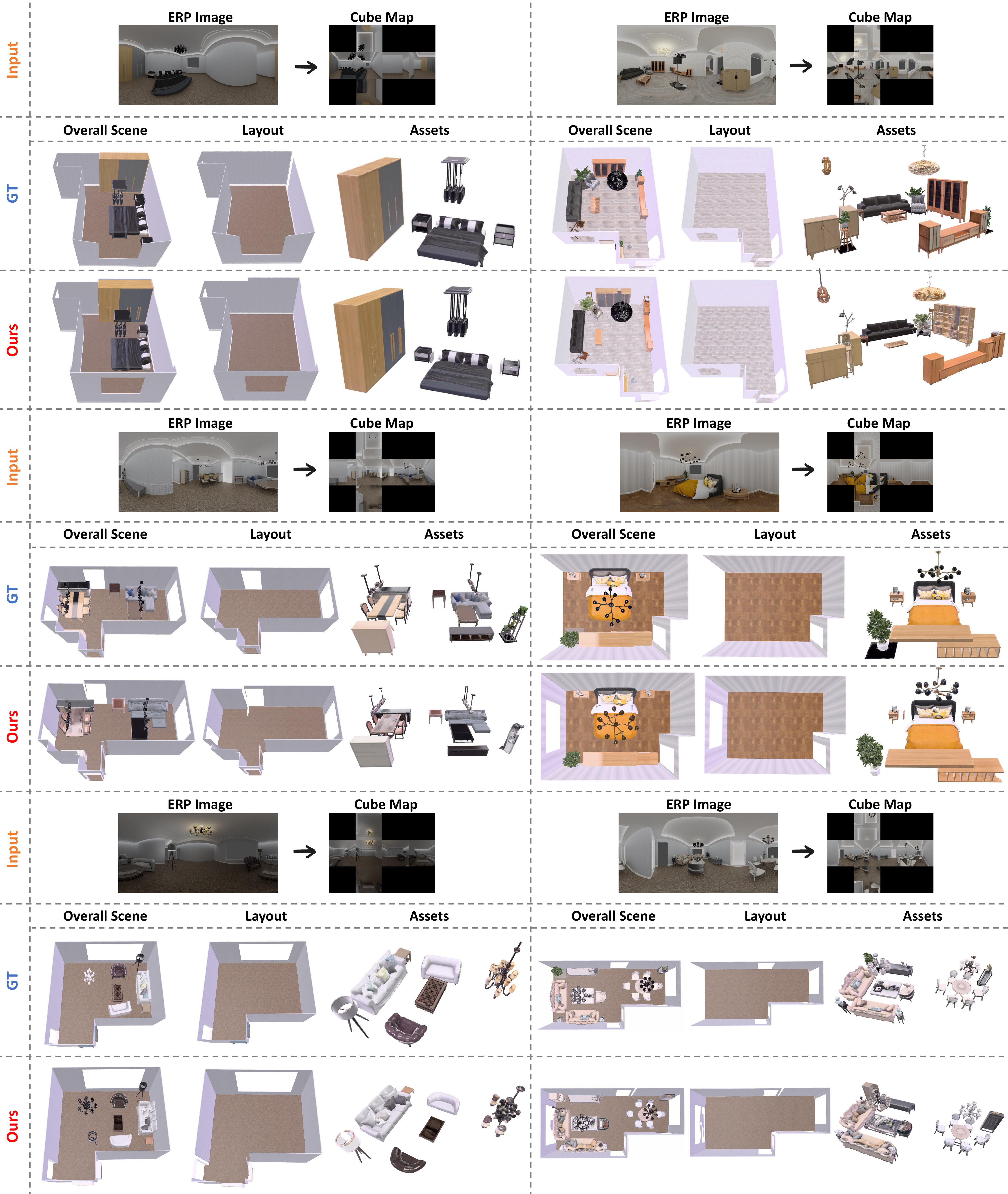}
   \caption{\textbf{Additional Qualitative Results on ERP-FRONT (Full Decomposition).} Given an ERP image and its cubemap decomposition (top), InSpace generates the overall scene, structural layout, and individual assets. Six examples are shown with ground truth (GT) for comparison.}
\label{fig:supple_5}
\end{figure}

\begin{figure}[p]
\centering
    \includegraphics[width=1.0\textwidth]{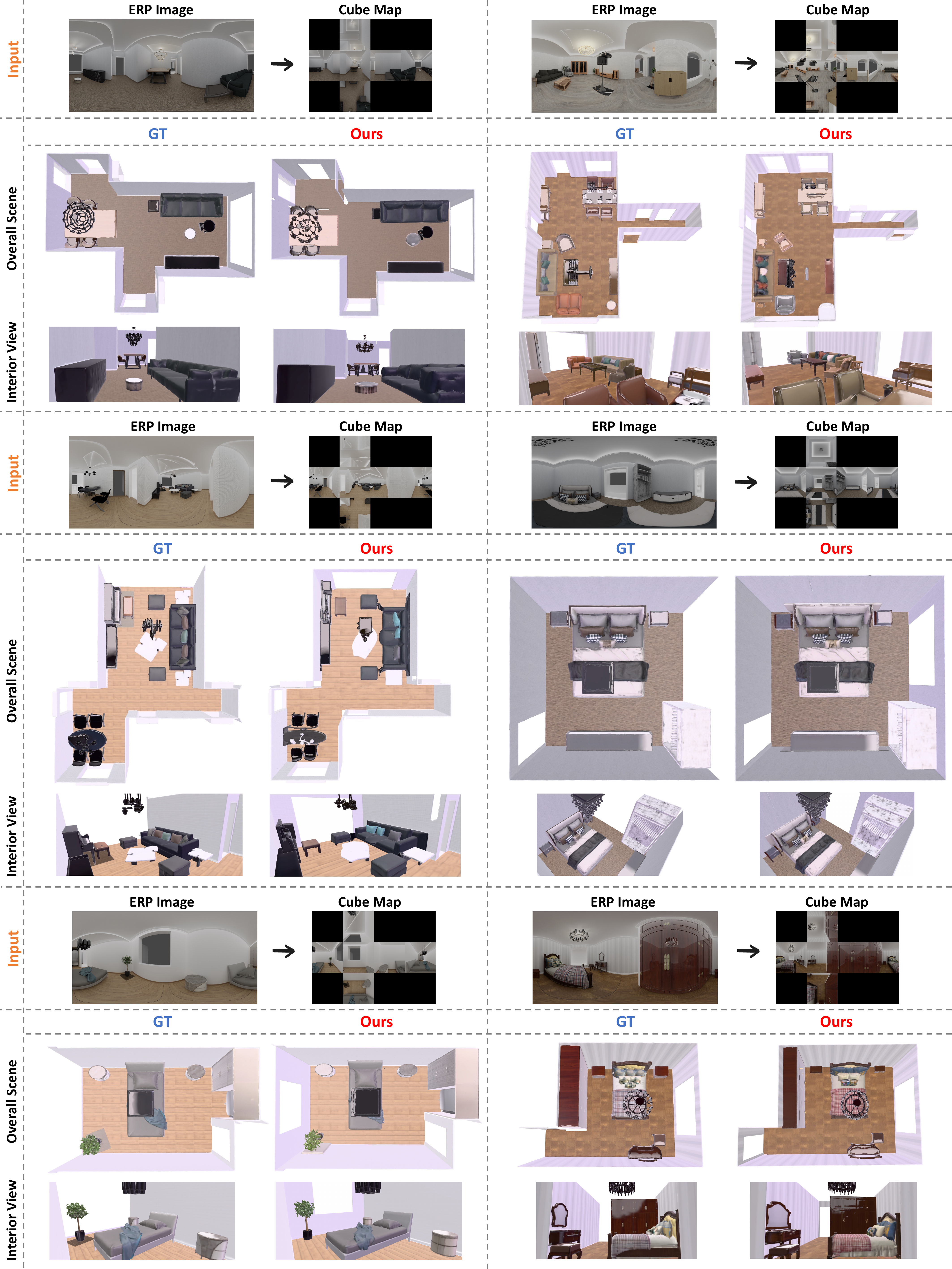}
   \caption{\textbf{Additional Qualitative Results on ERP-FRONT (Overall Scene and Interior View).} Given an ERP image and its cubemap decomposition (top), InSpace generates the complete 3D scene. For each example, we show a top-down overall scene view and an interior view rendered from the camera center, compared against ground truth (GT).}
\label{fig:supple_6}
\end{figure}

\begin{figure}[p]
\centering
    \includegraphics[width=1.0\textwidth]{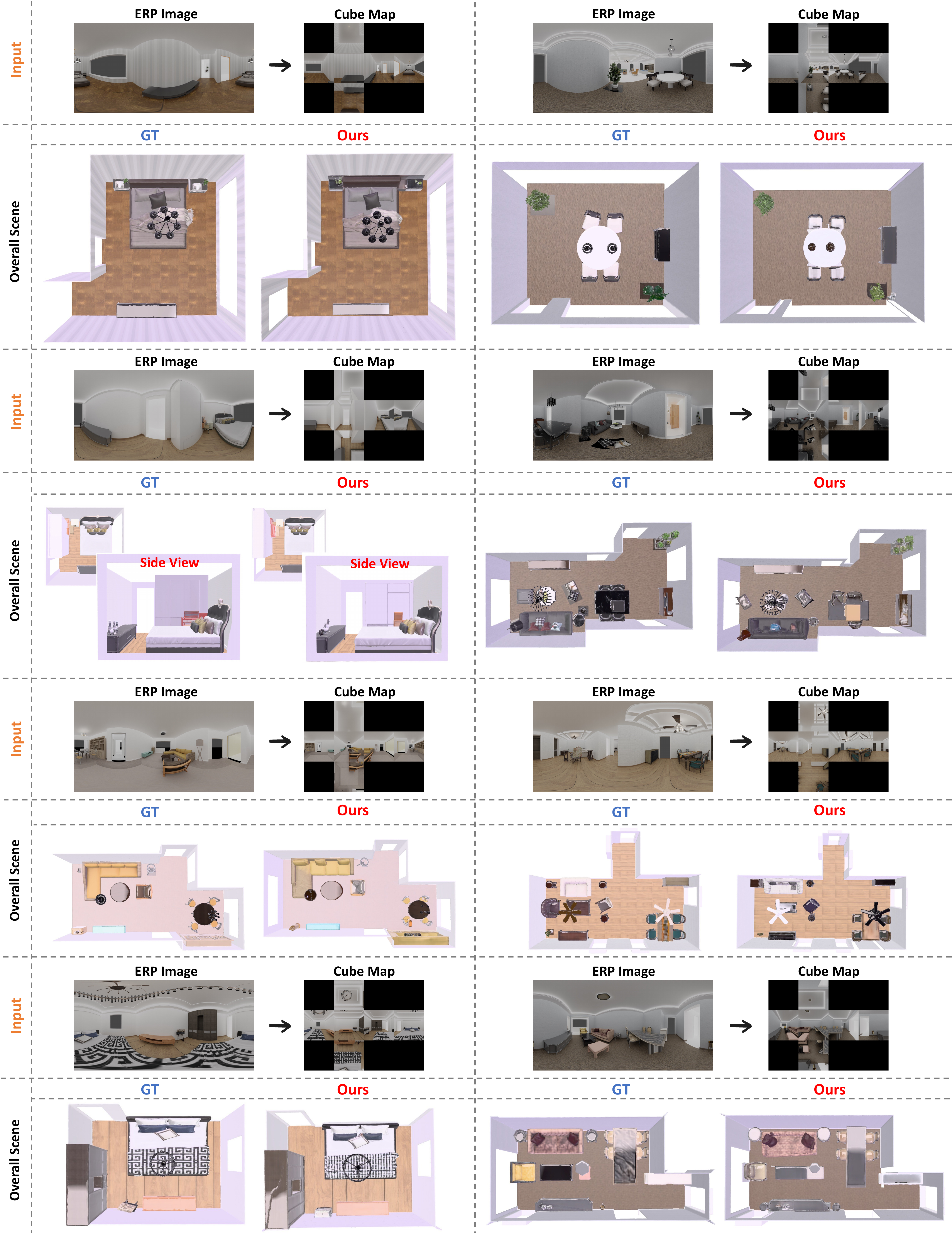}
   \caption{\textbf{Additional Qualitative Results on ERP-FRONT (Overall Scene).} Given an ERP image and its cubemap decomposition (top), we show the generated overall scene compared against ground truth (GT) across a diverse range of room types and layouts.}
\label{fig:supple_7}
\end{figure}

\section{More Qualitative Results}
\label{sec:supple_3_qualitative}

\subsection{Qualitative Results on ERP-FRONT}
We present additional qualitative results on the ERP-FRONT test set, complementing Fig. 8 of the main paper. Fig.~\ref{fig:supple_5} shows the complete decomposition of each generated scene into Overall Scene, Layout, and individual Assets. Fig.~\ref{fig:supple_6} provides Overall Scene alongside interior views rendered from the camera center. 
Fig.~\ref{fig:supple_7} presents additional Overall Scene samples to cover a broader range of room types.
At inference, a straightforward implementation of the scene decomposition in Stage 3 assigns voxels inside each 3D OBB exclusively to the corresponding asset, leaving the remainder as layout. This causes structural surfaces (floors and walls) overlapping with asset bounding boxes to be partially stripped from the layout. To address this, we introduce a structure-preserving decomposition strategy: voxels corresponding to floor and wall surfaces are identified based on their 3D coordinates and included in both the asset and layout components when they fall within an asset's OBB. This preserves structural surfaces in the layout regardless of asset placement, enabling clean separation between the layout and assets. As a result, Figs.~\ref{fig:supple_5}--\ref{fig:supple_7} demonstrate clean and complete layout--asset decomposition throughout.

\subsection{Qualitative Results on ReplicaPano}
\begin{figure}[p]
\centering
    \includegraphics[width=0.9\textwidth]{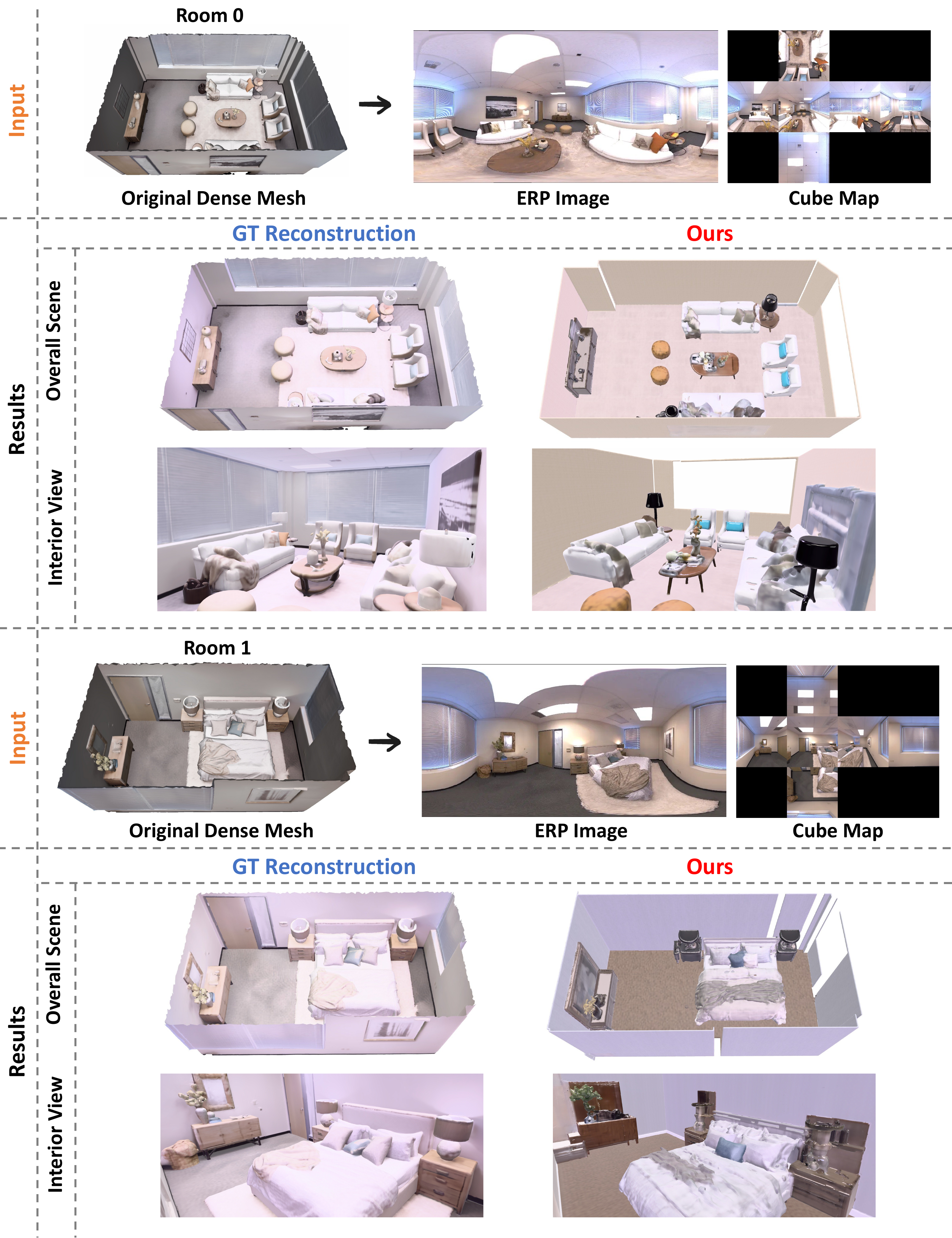}
   \caption{\textbf{Qualitative Results on ReplicaPano (Room~0, Room~1).}Given the original dense mesh and a rendered ERP image as input (top), InSpace generates a complete 3D indoor scene. We show the overall scene (top-down view) and an interior view rendered from the camera center, compared against the GT reconstruction. InSpace produces spatially coherent scenes despite the domain gap between synthetic training data and real-world inputs.}
\label{fig:supple_8}
\end{figure}

\begin{figure}[p]
\centering
    \includegraphics[width=0.9\textwidth]{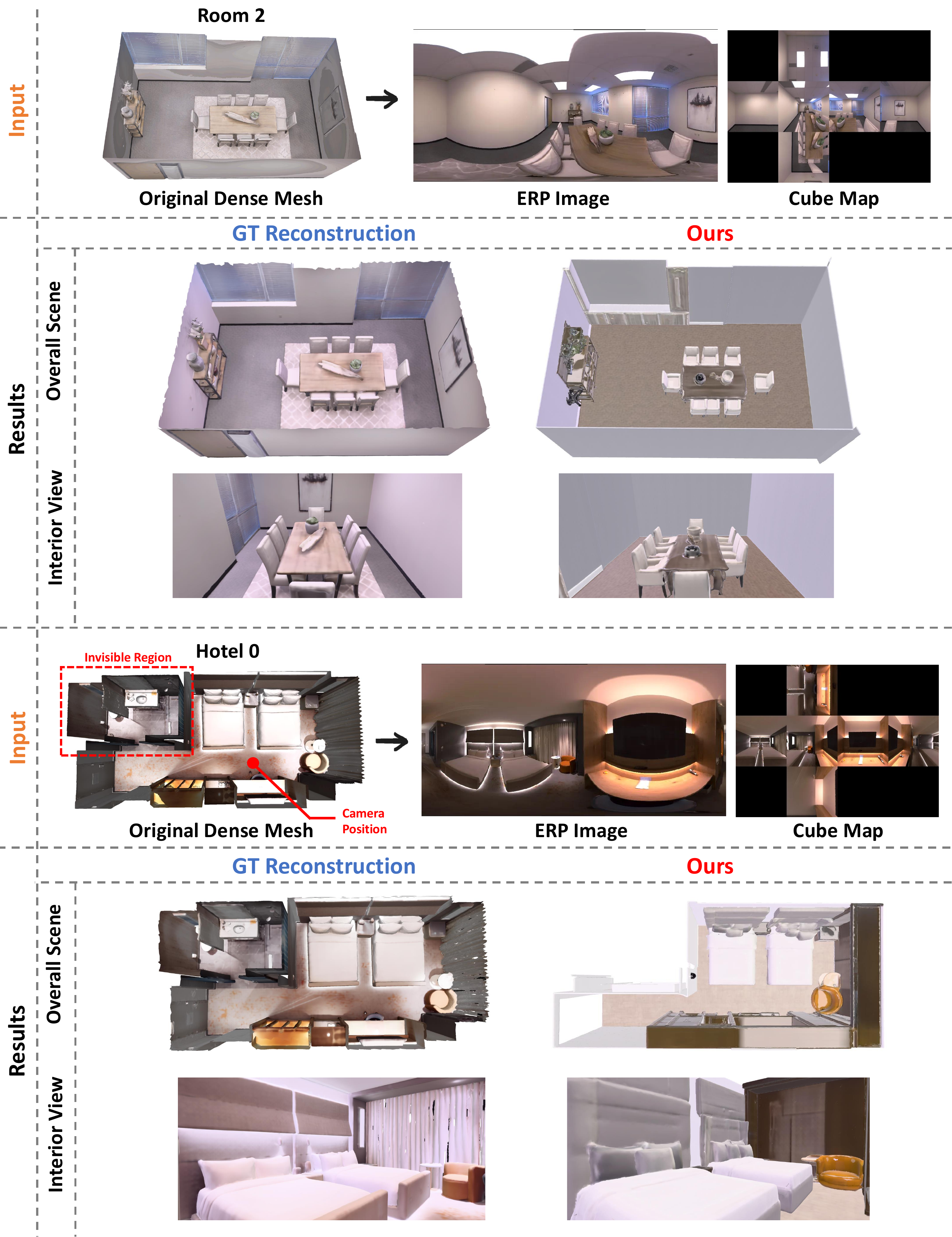}
   \caption{\textbf{Qualitative Results on ReplicaPano (Room~2, Hotel~0).}Given the original dense mesh and a rendered ERP image as input (top), InSpace generates a complete 3D indoor scene. For Hotel~0, the red dashed box indicates the invisible region due to camera placement, which cannot be reconstructed from the input ERP image. Overall scene (top-down view) and interior view are shown alongside the GT reconstruction.}
\label{fig:supple_9}
\end{figure}

To evaluate InSpace on realistic data beyond the synthetic ERP-FRONT dataset, we test on ReplicaPano~\cite{dong2024panocontext}, which provides ERP images extracted from the Replica dataset~\cite{straub2019replica}. Replica contains 18 high-quality indoor scenes in total; we exclude apartment 0, apartment 1, and apartment 2 as their scene meshes consist of multiple connected rooms, making single-room evaluation infeasible, and evaluate on the remaining 7 indoor scenes. For quantitative evaluation, we use the SC-VAE~\cite{xiang2025native} reconstruction of each scene mesh as the GT reference, as it shares the same coordinate system as our generated results, enabling reliable ICP alignment for direct metric computation. Results for Room~0 and Room~1 are shown in Fig.~\ref{fig:supple_8}, and Room~2 and Hotel~0 in Fig.~\ref{fig:supple_9}. 

As shown in Figs.~\ref{fig:supple_8}--\ref{fig:supple_9}, InSpace generalizes to real-world panoramic inputs despite being trained solely on synthetic data. The overall room layout and dominant furniture are recovered with plausible geometry and texture. However, as expected from the domain gap, the generated results deviate more from the reference compared to ERP-FRONT. Objects that are uncommon in the training distribution, such as wall-mounted frames and curtains, are not well reconstructed. Textures also reflect the training domain, tending toward cleaner and more uniform appearances than the real-world reference. 

\subsection{Quantitative Results on ReplicaPano}
Table~\ref{tab:replicapano} shows scene-level 3D and 2D reconstruction metrics on 7 ReplicaPano scenes. For 3D evaluation, we measure Chamfer Distance (CD), F-score at three thresholds ($\tau \in \{0.01, 0.05, 0.10\}$), and volumetric IoU. For 2D evaluation, we render 8 views per scene and measure PSNR, SSIM, LPIPS, CLIP similarity, and DINOv2 similarity against the reference mesh renders. Despite the domain gap between synthetic training data and real-world scans, InSpace achieves strong volumetric overlap (Vol.~IoU: 87.3\%) and high F-score at $\tau{=}0.10$ (93.7\%), while 2D metrics confirm visually consistent appearance (PSNR: 20.84\,dB, SSIM: 0.922).

\begin{table}[h]
\centering
\caption{\textbf{Quantitative Results on ReplicaPano.} Per-scene and 
mean 3D and 2D reconstruction metrics. CD is scaled by $\times 10^{-3}$.}
\label{tab:replicapano}
\resizebox{\linewidth}{!}{%
\begin{tabular}{lcccccc|ccccc}
\toprule
\multirow{2}{*}{Scene} 
  & \multicolumn{6}{c|}{3D Metrics} 
  & \multicolumn{5}{c}{2D Metrics} \\
\cmidrule(lr){2-7}\cmidrule(lr){8-12}
  & CD$\downarrow$ & CD$_{\text{p}{\to}\text{g}}$$\downarrow$ & CD$_{\text{g}{\to}\text{p}}$$\downarrow$
  & F@0.01$\uparrow$ & F@0.05$\uparrow$ & Vol.IoU$\uparrow$
  & PSNR$\uparrow$ & SSIM$\uparrow$ & LPIPS$\downarrow$ & CLIP$\uparrow$ & DINO$\uparrow$ \\
\midrule
frl\_apt\_2  & 4.44 & 1.30 & 3.14 & 24.5 & 76.6 & 83.9 & 19.42 & 0.928 & 0.105 & 0.880 & 0.611 \\
frl\_apt\_4  & 4.64 & 2.56 & 2.07 & 22.9 & 83.5 & 91.5 & 20.39 & 0.924 & 0.100 & 0.888 & 0.568 \\
hotel\_0     & 11.01 & 0.91 & 10.10 & 24.0 & 80.4 & 94.0 & 18.52 & 0.911 & 0.122 & 0.899 & 0.483 \\
office\_0    & 3.84 & 1.15 & 2.69 & 31.6 & 83.6 & 89.3 & 19.01 & 0.887 & 0.134 & 0.870 & 0.293 \\
room\_0      & 9.91 & 0.56 & 9.36 & 21.3 & 87.5 & 86.7 & 21.80 & 0.936 & 0.113 & 0.874 & 0.310 \\
room\_1      & 8.82 & 1.23 & 7.59 & 22.6 & 76.9 & 81.5 & 22.14 & 0.927 & 0.111 & 0.894 & 0.375 \\
room\_2      & 0.71 & 0.39 & 0.32 & 36.4 & 98.2 & 84.6 & 24.62 & 0.943 & 0.100 & 0.899 & 0.635 \\
\midrule
\textbf{Mean} 
  & \textbf{6.19} & \textbf{1.16} & \textbf{5.04} 
  & \textbf{26.2} & \textbf{83.8} & \textbf{87.3} 
  & \textbf{20.84} & \textbf{0.922} & \textbf{0.112} & \textbf{0.886} & \textbf{0.468} \\
\bottomrule
\end{tabular}%
}
\end{table}

\section{Comparison with Single-Image to Scene Generation Methods}
\label{sec:supple_4_comparison}

\subsection{Qualitative Results}

Since no existing method directly addresses ERP-image-to-scene generation, we compare InSpace against recent single-image scene generation methods: MIDI~\cite{huang2025midi}, SceneGen~\cite{meng2026scenegen}, and SAM3D~\cite{chen2026sam}. As these methods take a perspective image and its segmentation mask as input, we extract a perspective view from each ERP image by selecting the horizontal crop that maximizes the number of non-truncated assets, based on back-projecting 3D bounding box information. This perspective image is provided to the baseline methods, while InSpace receives the full ERP image. We conduct this comparison on 20 test scenes.
Qualitative results are shown in Fig.~\ref{fig:supple_10}. As the baseline methods are designed for perspective images where all assets are fully visible, they struggle with the partial visibility and occlusions present in our extracted views. MIDI and SceneGen frequently fail to generate complete or well-structured scenes under these conditions. SAM3D is more robust to occlusion due to its training strategy, and produces relatively better results. Nevertheless, InSpace consistently generates complete scenes with both structural layout and well-localized assets, benefiting from the full 360° context of the ERP image and depth-guided spatial grounding. The red dashed boxes in GT and InSpace (Ours) indicate the region used for asset-level quantitative evaluation, as detailed in Sec.~\ref{sec:supple_4_quan}.

\subsection{Quantitative Results}
\label{sec:supple_4_quan}
\begin{figure}[p]
\centering
    \includegraphics[width=0.9\textwidth]{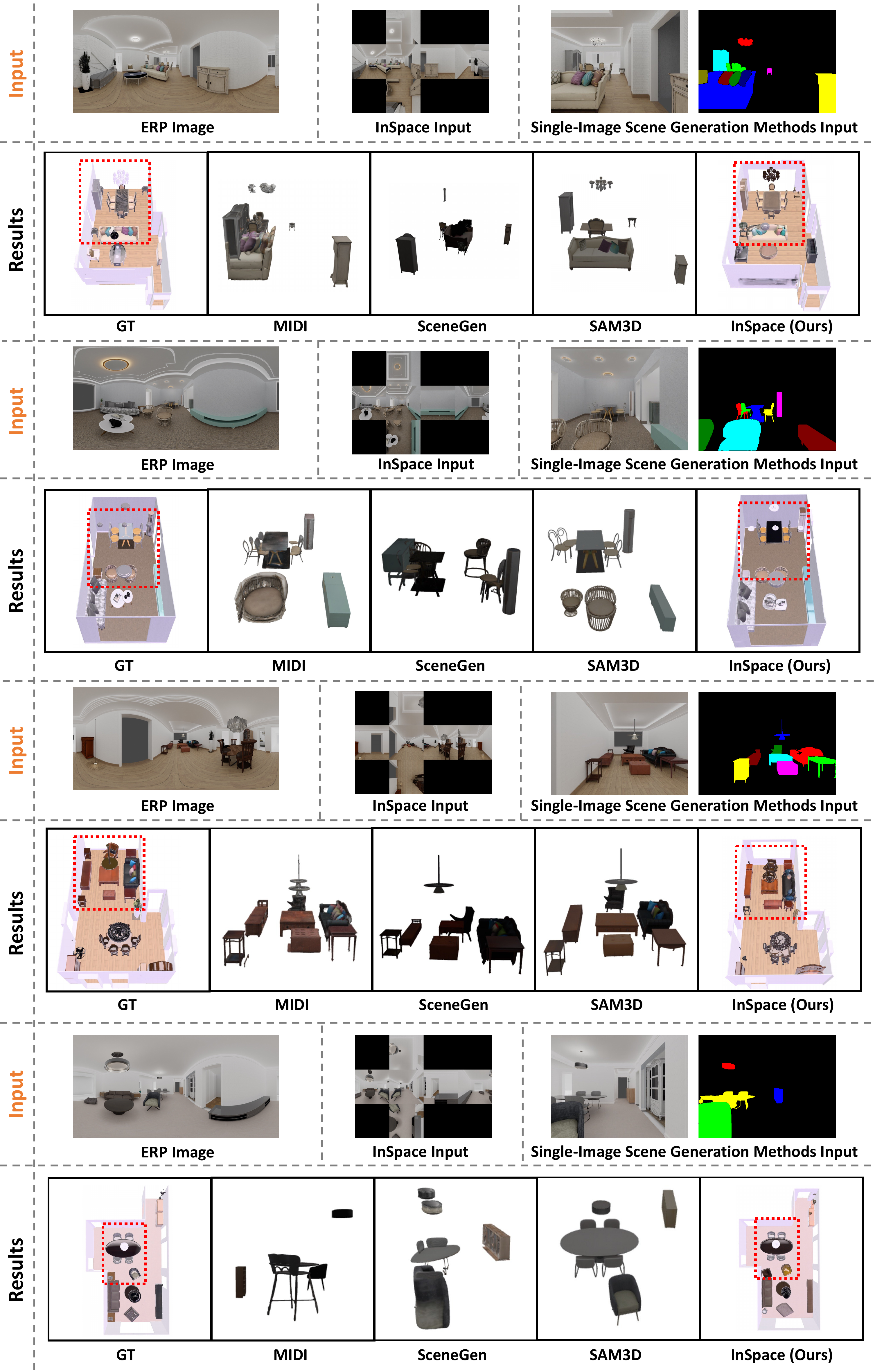}
   \caption{\textbf{Qualitative Comparison with Single-Image Scene Generation Methods.} For each scene, the ERP image, extracted perspective image, and segmentation mask are shown at the top. MIDI and SceneGen frequently fail under partial visibility, while SAM3D is more robust to occlusion. InSpace consistently produces spatially coherent scenes with accurate asset placement by leveraging full 360° context and depth-guided grounding. Red dashed boxes indicate the region used for quantitative evaluation in Table~\ref{tab:comparison_asset}.}
\label{fig:supple_10}
\end{figure}

Table~\ref{tab:comparison_asset} reports asset-level 3D metrics on the 20 test scenes, evaluated only within the perspective region indicated by the red dashed boxes in Fig.~\ref{fig:supple_10}. MIDI and SceneGen show significantly higher CD and lower F-scores, consistent with their qualitative failure to reconstruct assets under partial visibility. SAM3D performs substantially better owing to its occlusion-robust training, yet InSpace achieves the best results across all metrics, demonstrating that depth-guided spatial grounding from the full 360° ERP image leads to more accurate asset-level reconstruction than perspective-based approaches.

\begin{table}[t!]
\centering
\caption{\textbf{Asset-Level Quantitative Comparison with Single-Image 
Scene Generation Methods.} CD is scaled by $\times 10^{-3}$.}
\label{tab:comparison_asset}
\begin{tabular}{lccccc}
\toprule
Method & CD$\downarrow$ & CD$_{\text{p}{\to}\text{g}}$$\downarrow$ & CD$_{\text{g}{\to}\text{p}}$$\downarrow$ & F@0.01$\uparrow$ & F@0.05$\uparrow$ \\
\midrule
MIDI~\cite{huang2025midi}                          & 23.5          & 4.4          & 19.1          & 16.8          & 55.4          \\
SceneGen~\cite{meng2026scenegen} & 24.2          & 4.5          & 19.7          & 14.2          & 46.9          \\
SAM3D~\cite{chen2026sam}          & 6.7           & 1.3          & 5.4           & 23.3          & 76.9          \\
\textbf{InSpace (Ours)}                            & \textbf{6.2}  & \textbf{1.2} & \textbf{5.0}  & \textbf{24.2} & \textbf{79.9} \\
\bottomrule
\end{tabular}
\end{table}

\section{Limitation and Future Work}
\label{sec:supple_6_limitation}

\noindent\textbf{Limitation.} While InSpace demonstrates strong performance on synthetic indoor scenes, several limitations remain. First, as shown in the ReplicaPano experiments, a domain gap exists between the synthetic training data (ERP-FRONT) and real-world panoramic scans. Objects common in real environments but underrepresented in 3D-FRONT, such as wall-mounted frames, curtains, and irregular furniture, are not well reconstructed, and generated textures tend to reflect the synthetic training distribution. Second, the sparse voxel latent representation operates at a fixed resolution of $64^3$, which limits geometric detail in large rooms where fine-grained structure must be captured within a coarse grid. Third, since InSpace relies on a single 360° ERP image, regions that are heavily occluded or outside the observable field of view from the camera position may be reconstructed with reduced fidelity. Finally, the asset generation is constrained to object categories present in the 3D-FRONT training distribution, limiting generalization to out-of-distribution room types or furniture styles.

\noindent\textbf{Future Work.} Several directions are promising for addressing these limitations. Incorporating real-world panoramic data during training or leveraging domain adaptation techniques could reduce the synthetic-to-real gap and improve generalization to in-the-wild inputs. Adopting hierarchical or adaptive voxel representations could preserve fine-grained geometry in large-scale scenes without a fixed resolution bottleneck. Finally, extending the input modality to multi-view panoramic images or video sequences would provide richer spatial context, enabling more complete and accurate reconstruction of occluded regions and large-scale environments. 
Beyond reconstruction, the explicit decomposition of a scene into structural layout and individual assets may facilitate future research on controllable scene understanding and manipulation. Asset-level grounding could enable localized modifications while preserving global scene consistency, opening opportunities for interactive indoor scene editing.
Another promising direction is to incorporate user guidance into the generation process. Allowing users to iteratively refine layouts, assets, or scene appearance could improve controllability and make generated environments more practical for real-world content creation workflows \cite{koo2024flexiedit,koo2025flowdrag,yoon2025occlusion,yoon2024frag,yoon2024dni,yoon2024tpc,hong2025ita,koo2024wavelet}.